\documentclass[conference]{IEEEtran}
\usepackage{cite}
\usepackage{amsmath,amssymb,amsfonts}
\usepackage{algorithmic}
\usepackage{graphicx}
\usepackage{textcomp}
\usepackage{xcolor}
\def\BibTeX{{\rm B\kern-.05em{\sc i\kern-.025em b}\kern-.08em
    T\kern-.1667em\lower.7ex\hbox{E}\kern-.125emX}}

\usepackage{latexsym}

\usepackage{url}            
\usepackage{booktabs}       
\usepackage{amsfonts}       
\usepackage{nicefrac}       
\usepackage{microtype}      
\usepackage{bm}
\usepackage{amsmath}
\usepackage{amssymb}
\usepackage{amsthm}
\usepackage[colorlinks,allcolors=black]{hyperref}
\newcommand{\abs}[1]{\ensuremath{\lvert#1\rvert}}
\renewcommand{\vec}[1]{\ensuremath{\mathbf{#1}}}
\usepackage{xcolor}
\usepackage{mathtools}
\usepackage{xspace}
\usepackage{graphicx}
\usepackage{siunitx}
\usepackage{enumitem}
\usepackage{afterpage}
\usepackage{subcaption}
\usepackage{dsfont}
\usepackage[algo2e,ruled,vlined,linesnumbered]{algorithm2e}
\SetKwComment{Comment}{$\triangleright$\ }{}
\usepackage{float}

\newcommand{\W}{\ensuremath{\mathbf{W}}\xspace}
\newcommand{\fs}{\ensuremath{f^{\text{S}}}\xspace}
\newcommand{\ft}{\ensuremath{f^{\text{T}}}\xspace}
\newcommand{\fhats}{\ensuremath{\hat{f}^{\text{S}}}\xspace}
\newcommand{\fhatt}{\ensuremath{\hat{f}^{\mkern1mu\text{T}}}\xspace}
\newcommand{\SO}[1]{\ensuremath{\operatorname{SO}\left(#1\right)}\xspace}

\usepackage[ruled,vlined]{algorithm2e} 

\usepackage{newrevisor}
\newrevisor{manuel}{violet!75}

\begin{document}

\title{Transfer Learning of Surrogate Models via Domain Affine Transformation Across Synthetic and Real-World Benchmarks
}

\author{
    Shuaiqun Pan\textsuperscript{1}, 
    Diederick Vermetten\textsuperscript{1}, 
    Manuel López-Ibáñez\textsuperscript{2}, 
    Thomas Bäck\textsuperscript{1}, 
    Hao Wang\textsuperscript{1} \\[5pt]
    \textsuperscript{1}\textit{LIACS, Leiden University}, Leiden, The Netherlands \\
    \textsuperscript{2}\textit{University of Manchester}, Manchester, UK \\[5pt]
    \{s.pan, d.l.vermetten, t.h.w.baeck, h.wang\}@liacs.leidenuniv.nl, manuel.lopez-ibanez@manchester.ac.uk
}


\IEEEoverridecommandlockouts
\IEEEpubid{\makebox[\columnwidth]{ 979-8-3315-3431-8/25/\$31.00~\copyright2025 IEEE \hfill} 
\hspace{\columnsep}\makebox[\columnwidth]{ }}


\maketitle
\IEEEpubidadjcol
\begin{abstract}
Surrogate models are frequently employed as efficient substitutes for the costly execution of real-world processes. However, constructing a high-quality surrogate model often demands extensive data acquisition. A solution to this issue is to transfer pre-trained surrogate models for new tasks, provided that certain invariances exist between tasks. This study focuses on transferring non-differentiable surrogate models (e.g., random forests) from a source function to a target function, where we assume their domains are related by an unknown affine transformation, using only a limited amount of transfer data points evaluated on the target. Previous research attempts to tackle this challenge for differentiable models, e.g., Gaussian process regression, which minimizes the empirical loss on the transfer data by tuning the affine transformations. In this paper, we extend the previous work to the random forest and assess its effectiveness on a widely-used artificial problem set - Black-Box Optimization Benchmark (BBOB) testbed, and on four real-world transfer learning problems. The results highlight the significant practical advantages of the proposed method, particularly in reducing both the data requirements and computational costs of training surrogate models for complex real-world scenarios.
\end{abstract}


\section{Introduction}
Surrogate modeling~\cite{DBLP:books/daglib/0022623, forrester2009recent, DBLP:journals/cce/BhosekarI18, DBLP:journals/isci/TongHMY21} is commonly utilized across various fields of computer science and engineering, particularly for complex simulations or physical experiments where data acquisition is expensive. 
Common learning algorithms for building surrogate models include artificial neural networks~\cite{DBLP:conf/gecco/DushatskiyMAB19}, support vector machine~\cite{DBLP:journals/tec/NguyenXZ24, DBLP:journals/ress/ChengL21}, random forests~\cite{DBLP:journals/tcyb/WangJ20, DBLP:journals/ress/AntoniadisLP21} and Gaussian process regression~\cite{satria2020gaussian, rajaram2020deep, DBLP:journals/technometrics/Pourmohamad21}. These learning algorithms often have a high sample complexity, especially when the dimension of the independent variable becomes large. As such, when surrogate-modeling a new regression task, one might wish to exploit surrogate models trained on previous tasks, given similarities and symmetries between tasks.

Domain shift is a prominent type of task similarity, where the probability distribution of the input variable shifts from one task to the other while the predictive distribution remains the same. Under this assumption, one can transfer a model from the source task to the target task by learning the domain shift. 
In this work, we focus on a specific type of domain shift - affine shift - the input distributions between tasks are related by an unknown affine transformation (rotation and translation).  
A prior study~\cite{DBLP:conf/gecco/PanV0B024} proposed a transfer learning method for differential models, e.g., Gaussian process regression. In this work, we propose to extend this transfer learning method to non-differentiable models, for instance, random forests. 
\begin{figure*}[!ht]
\centering
\begin{tabular}{*{7}{>{\centering\arraybackslash\small}p{.118\textwidth}}}
Source function & Target function & Original RFR & Trained with\newline 50 samples & Transferred with 50 samples & Trained with\newline 100 samples & Transferred with 100 samples \\
\end{tabular}
\includegraphics[width=\textwidth]{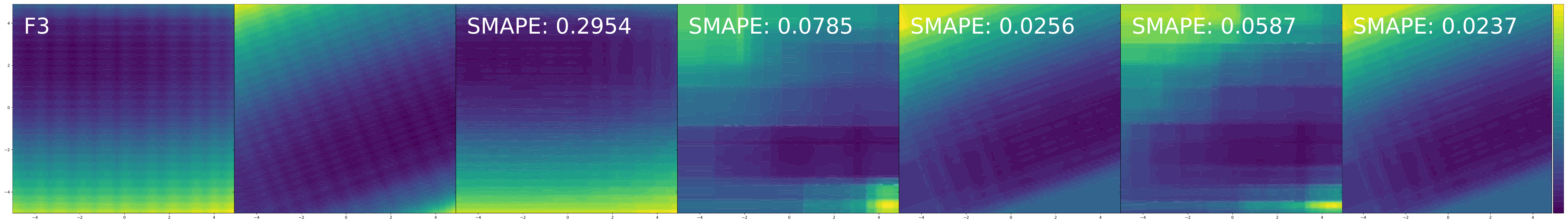}
\caption{
Synthetic transfer learning tasks are constructed by transferring from different instances of the same BBOB functions (F3 Rastrigin is taken in this example).
From left to right, we show the \emph{source function} \fs, the \emph{target function} \ft, the \emph{original RFR} \fhats trained to approximate \fs, the RFR trained directly on 50 samples of \ft, and the original RFR transferred with the same 50 samples. We also show the RFR models with 100 sample points.
}
\label{figure:transferRF_BBOB}
\end{figure*}

We illustrate our approach in Fig.~\ref{figure:transferRF_BBOB} with a random forest regression (RFR) model on an artificial function taken from BBOB. The first three columns show the contour lines of the source function, the target function, and the model \fhats, respectively. The next two columns show the contour lines of the RFR model trained from scratch with 50 sample points evaluated on $\ft$ and the transferred model. The last two columns display the situation with 100 sample points. 
As the target function is essentially a ``rotated'' variant of the source function, the transferred model demonstrates a significant advantage over the one trained from scratch, achieving markedly lower symmetric mean absolute percentage error (SMAPE).

Our contributions are:
\begin{itemize}
    \item We extend the transfer learning method in~\cite{DBLP:conf/gecco/PanV0B024} to handle non-differentiable surrogates, e.g., random forest regression.
    \item We verify the effectiveness of our extension on the Black-Box Optimization Benchmark suite and four real-world transfer learning problems.
    \item We conduct an empirical analysis to investigate when the transferred model outperforms the model trained from scratch on the transfer dataset.
\end{itemize}

\section{Related Works}
\subsection{Transfer learning in RFR surrogate modeling}
Segev et al.~\cite{DBLP:journals/pami/SegevHMCE17} introduced two distinct algorithms for transferring RFR. The first method employs a greedy search strategy to iteratively refine the structure of each decision tree. This approach enables localized modifications, such as expanding or pruning individual nodes, to improve the model's adaptability. The second method, in contrast, preserves the original tree structure but focuses on recalibrating the parameters, specifically by adjusting the decision thresholds at the internal nodes.

\subsection{Transfer learning with affine model}
Saralajew et al.~\cite{saralajew2017transfer} proposed a manifold-based adaptation technique for classification models, such as generalized learning vector quantization classifiers (GLVQ)~\cite{kohonen1990improved, kaden2014aspects}, to address distribution shifts between datasets. The affine transformation was also considered to minimize discrepancies between source and target domains~\cite{chen2019pose}.
Mandl et al.~\cite{mandl2023affine} applied affine transformations within the framework of Physics-Informed Neural Networks (PINNs).

\newcommand{\ntransfer}{\ensuremath{n_\text{T}}}

\section{Affine Transfer Learning for Non-differentiable Models}\label{sec:method}
Consider a source regression task to approximate function $\fs\colon \mathbb{R}^d\to \mathbb{R}$ and a target task $\ft\colon \mathbb{R}^d\to \mathbb{R}$.
We assume an affine symmetry between them: there exist $\W\in\SO{d}$ and $\mathbf{v}\in\mathbb{R}^d$ such that $\ft(\mathbf{x}) = \fs(\W\mathbf{x} + \mathbf{v})$, where $\SO{d}$ denotes the $d$-dimensional rotation group. In practice, the affine parameters \W and $\mathbf{v}$ are unknown. Given a high-quality model $\fhats$ trained on $\fs$, we wish to transfer it to $\ft$ by affine reparameterization $\fhatt(\vec{x}) \coloneqq \fhats(\W\mathbf{x} + \mathbf{v})$, where the unknown parameters are determined by minimizing the empirical loss of $\fhatt$ on a small transfer data set $\mathcal{T}$ evaluated on $\ft$, namely $\mathcal{L}(\mathbf{v}, \W) =\ntransfer^{-1}\sum_{\mathbf{x}\in \mathcal{T}}(\fhats\left(\W\mathbf{x} + \mathbf{v}\right) - \ft(\mathbf{x}))^2$.
If the surrogate model $\fhats$ is differentiable, e.g., GPR, we can employ the Riemannian gradient algorithm to minimize the loss~\cite{DBLP:conf/gecco/PanV0B024}. For non-differential models, e.g., RFR, we propose the following optimization procedure.

We propose to use Covariance matrix adaptation evolution strategy (CMA-ES)~\cite{DBLP:reference/sp/EmmerichS018,DBLP:books/sp/06/Hansen06, DBLP:journals/corr/Hansen16a} to tune the affine parameters. CMA-ES can be applied directly to the search space of the translation parameter $\mathbf{v}$, which is Euclidean. However, special treatment is needed for \W, which lives in a smooth manifold \SO{d}. To solve this issue, we consider the Lie group representation $\mathfrak{so}(d)=\{\mathbf{A}\in\mathbb{R}^{d\times d}\colon \mathbf{A}^\top = -\mathbf{A}\}$, which is a flat space (with dimension $d(d-1)/2$), and optimize this representation with CMA-ES. A rotation matrix \W can be recovered its representation $\mathbf{A}$ with the exponential map, i.e., $\W = \operatorname{Exp}(\mathbf{A})$. For each search point $\mathbf{z}\in\mathbb{R}^{d(d-1)/2}$, we have to transform it into a $d\times d$ antisymmetric matrix to preserve the structure of $\mathfrak{so}(d)$: the components in $\mathbf{z}$ are sequentially assigned to the upper triangular of $\mathbf{A}$ (diagonal entries are zero) row by row. The lower triangular entries are then filled by the negative value of the transposition of the upper triangular. We summarize this Riemannian version of CMA-ES in Alg.~\ref{alg:cmaes}.

\begin{algorithm2e}[t]
\caption{Transfer Learning via Learning Affine Transformation with CMA-ES}\label{alg:cmaes}
\textbf{Procedure:} TL-CMA-ES($\mathcal{L}$, $\lambda$, $\sigma$)\;
\textbf{Input:} the empirical loss $\mathcal{L}$, population size $\lambda$, the initial step-size $\sigma$, termination conditions\;
Sample $\mathbf{z}$ uniformly at random~in $\mathbb{R}^{d(d-1)/2}$ and $\mathbf{v}$ uniformly at random~in $\mathbb{R}^{d}$\;
$\mathbf{m} \leftarrow (\mathbf{v}, \mathbf{z})^\top, \, \mathbf{C} \leftarrow \mathbf{I}$\;
\Repeat{termination conditions are met}{
    Sample $\{\mathbf{x}^i\}_{i=1}^{\lambda}$ i.i.d. from $\mathcal{N}(\mathbf{m}, \sigma^2 \mathbf{C})$\;
    \tcp{\small function evaluation}
    \For{$i\in[1..\lambda]$}{
        $\mathbf{v}^i\leftarrow (x^i_1, \ldots, x^i_d)$\;
        \tcp{\small construct antisymmetric matrices}
        $\mathbf{A}^i\leftarrow {\small\texttt{antisymmetric}}(x^i_{d+1}, \ldots, x^i_{d(d+1)/2})$\;
        $\mathbf{W}^i \leftarrow \operatorname{Exp}(\mathbf{A}^i)$\Comment*[r]{\small matrix exponential}
        $y^i\leftarrow \mathcal{L}(\mathbf{v}^i, \mathbf{W}^i)$\;
    }
    \tcp{\small CMA-ES parameter update~\cite{DBLP:journals/corr/Hansen16a}}
    Update $\mathbf{m},\mathbf{C}$ and $\sigma$ with and $\{\mathbf{x}^i\}_{i=1}^{\lambda}$ and $\{y_i\}_{i=1}^\lambda$\;
}
Extract $\mathbf{v}^*, \mathbf{W}^*$ from the best solution $\mathbf{x}^*$\;
\textbf{Output:} $\mathbf{v}^*, \mathbf{W}^*$
\end{algorithm2e}

\section{Experimental Settings}\label{sec:experiment}

\subsection{Synthetic tasks based on BBOB}
We evaluate our transfer learning approach with RFR using the widely recognized BBOB problem set. While originally designed for benchmarking black-box optimization algorithms, BBOB is also widely utilized as a regression testbed~\cite{singh2018integration, DBLP:journals/tai/TianPZRTJ20, DBLP:conf/nips/ChenSLW0DKKDRPF22, DBLP:conf/emo/YangA23} due to its computational efficiency and the diverse properties of its functions. These include multi-modality, ill-conditioning, and irregularities that closely resemble the complexities of real-world regression tasks. Detailed information on BBOB is available in its documentation~\cite{DBLP:journals/oms/HansenARMTB21}. To construct target problems for each source problem in BBOB, we introduce diversity and complexity by randomly generating rotation matrices and translation vectors, creating a robust and challenging testbed for our method.

We train the original RFR model $\fhats$ on a dataset generated by uniformly sampling $1\,000 \times d$ points at random within the range $[-5,5]^d$ and evaluating them on $\fs$. To apply our proposed transfer learning approach, we generate a transfer dataset $\mathcal{T}$ by uniformly sampling $50 \times d$ points within $[-5,5]^d$ and evaluating them on the target function  $\ft$. For comparison, we also train an RFR model from scratch on the same transfer dataset $\mathcal{T}$ without leveraging any prior knowledge. Finally, to evaluate the accuracy of the various RFR models on $\ft$, we independently sample uniformly at random a test dataset of $1\,000 \times d$ points evaluated on $\ft$.

\subsection{Real-World benchmark}

\subsubsection{Porkchop Plot Benchmarks in Interplanetary Trajectory Optimization}
The energy required to travel between Earth and another planet in the solar system is a function of the relative positions and velocities of the planets, as well as the desired trip duration, i.e., of both the departure and arrival times. Calculating this energy involves solving Lambert's problem, a differential equation that determines the orbital trajectory~\cite{izzo2015revisiting}. The periodic motion of planets generates recurring patterns in the contour lines of this function around local minima. These contours resemble the shape of a porkchop slice, inspiring the term ``porkchop plots''~\cite{cianciolo2006mars, natureearthmars2013}. The analysis of porkchop plots is useful in the design of interplanetary trajectories. For benchmarking, we consider as the source function the energy required for an Earth-to-Mars mission within a specific launch window and within a maximum travel time of 14 years and the target function corresponds to a different launch window. In this way, we aim to transfer insights gained from the planning of an interplanetary mission during a specific launch window to plan similar missions for future windows. Besides the Earth-to-Mars mission, we also validate our approach on Earth-to-Venus (a maximum travel time of 12 years) and Mercury-to-Earth (a maximum travel time of 6 years) missions.

For validating our transfer learning method on this dataset, for example, the energy function of a specific Earth-to-Mars mission during a chosen launch window is treated as the source function $\fs$, while the corresponding energy functions for alternative launch windows serve as the target functions $\ft$. The surrogate model $\fhats$ is trained using $40\,000$ randomly selected points from $\fs$, while the full dataset is used as the test set for evaluating the SMAPE of the various RFR models on the target function. The transfer dataset $\mathcal{T}$ is sampled uniformly at random from the full dataset of the target function. The same transfer dataset is used to train an RFR model from scratch without leveraging any prior knowledge. We study the effect of varying the size of the transfer dataset, ranging from 10 to $1\,000$ points.

\subsubsection{Kinematics of a Robot Arm} This task involves predicting the feed-forward torques needed for the seven joints of the SARCOS robot arm~\cite{DBLP:books/lib/RasmussenW06} to execute a desired trajectory. The input data comprises 21 features, including joint positions, velocities, and accelerations. The target is to predict the torque value for a specific joint. $30\,000$ randomly sampled points from the source function are used to train the surrogate model $\fhats$. Similarly, up to $1\,000$ points, also selected at random, are drawn from the target function $\ft$ to create the transfer dataset $\mathcal{T}$. 

\subsubsection{Real-World Optimization Benchmark in Vehicle Dynamics} This dataset benchmarks optimization algorithms in automotive applications, focusing on minimizing braking distance. It features five vehicle configurations, each defined by distinct tire performance and vehicle load combinations. The parameter space comprises two ABS control variables, $x_1$ and $x_2$, spanning $10\,101$ discrete combinations~\cite{DBLP:conf/ijcci/ThomaserVBK23}. The surrogate model $\fhats$ for source functions is trained using the entire $\fs$ dataset, while up to $10\,101$ points from target $\ft$ form the transfer dataset $\mathcal{T}$.

\begin{figure*}[!ht]
  \centering
  \includegraphics[width=\textwidth]{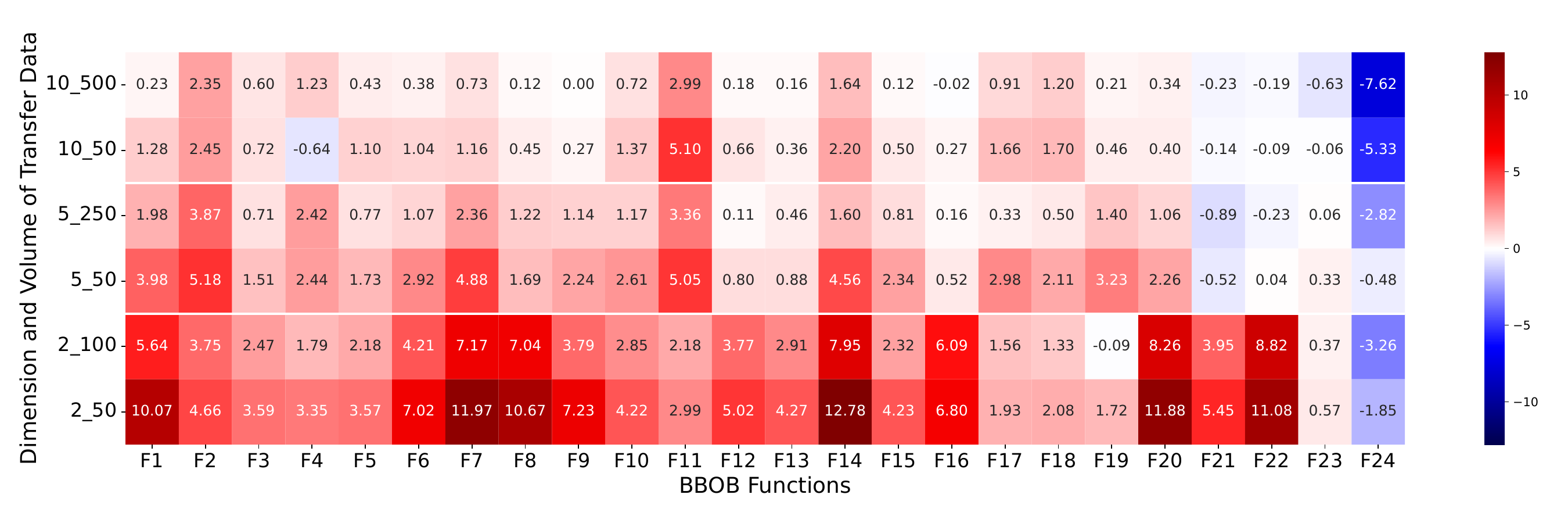}
  \caption{
  The comparison evaluates Random forest regression models obtained through transfer learning against those trained from scratch on the transfer dataset, across varying sample sizes and problem dimensions. Each cell in the figure shows the percentage difference in average SMAPE (\%) between the two approaches for specific BBOB functions. Positive values (marked in red) indicate better accuracy from transfer learning (i.e., lower SMAPE).}
  \label{figure:smape_heatmap}
\end{figure*}

\subsubsection{Single-Objective Game-Benchmark MarioGAN Suite} The RW-GAN-Mario suite~\cite{DBLP:conf/gecco/VolzNKT19} offers 28 single-objective fitness functions for evaluating and optimizing procedurally generated levels in the Mario AI environment. These functions cover various level types (e.g., underground) and ensure segment playability through concatenation mechanisms. Fitness metrics combine statistical properties (e.g., enemy distribution) with gameplay-based measures (e.g., air time) derived from simulations. In this work, we focus on three functions - F1, F5, and F6 - and apply transfer learning within instances of each function. For each source instance, the surrogate model $\fhats$ is trained on 50\,000 randomly sampled points, while up to $1\,000$ randomly selected points from the target instance $\ft$ form the transfer dataset $\mathcal{T}$.

\subsection{Surrogate model} We evaluate RFR using SMAPE~\cite{flores1986pragmatic} for its scale-invariant properties, allowing fair comparisons across datasets. Transfer learning experiments are repeated ten times with random affine transformations applied to each BBOB function. Similarly, real-world applications are also tested ten times to account for variability and ensure robust and consistent results.

\subsection{Implementation details}
Experiments are conducted using Python 3.9 using BBOB functions via the IOHexperimenter framework~\cite{DBLP:journals/ec/NobelYVWDB24}. RFR is implemented with \texttt{scikit-learn} library~\cite{scikit-learn} using default hyperparameters when validating on BBOB functions. CMA-ES is implemented with \texttt{pycma}~\cite{hansen2019pycma}, where the parameter vector encodes rotation and translation. Rotation is initialized by sampling a Gaussian antisymmetric matrix, applying the exponential map, and extracting the upper-triangular elements. Translation is randomly sampled from $[-0.5, 0.5]$ per dimension. Bounds are set to $[-\pi, \pi]$ for rotation and $[-1.5, 1.5]$ for translation, with the initial step-size set to one-fifth of the average range. All other settings follow \texttt{pycma} defaults, using BIPOP-CMA-ES with three restarts for robustness.

For the vehicle dynamics task, RFR hyperparameters are tuned using \texttt{SMAC3}~\cite{DBLP:journals/jmlr/LindauerEFBDBRS22}, optimizing tree count ($[100, 1\,500]$), depth ($[10, 60]$), split/leaf samples ($[2, 20]$, $[1, 10]$),  and features fractions ($[0.1, 1.0]$). Other real-world tasks use default settings with 500 trees. To address the skewed distribution of relative function values (differences from the global optimum) in BBOB functions, a $\ log$ transformation is applied before model fitting. For the Robot Arm Kinematics task, we follow the pre-processing~\cite{DBLP:conf/nips/MinamiFHY23}. Porkchop Plot benchmarks are generated using \texttt{poliastro}~\cite{canopoliastro}, with other data sourced from their respective references.


\subsection{Reproducibility}
The complete experimental setup, implementation, and supplementary materials are available in our Zenodo repository~\cite{Anonymous-supp} for full reproducibility.

\section{Experimental Results}

\begin{figure*}[!ht]
  \centering
  \includegraphics[width=\textwidth]{Plots/smape_heatmap_Real_world_Porkchop_plot_Earth_Mars.pdf}
  \caption{The comparison evaluates Random forest regression models obtained through transfer learning against those trained from scratch on Porkchop Plot Benchmarks for Earth-to-Mars trajectory optimization. It examines different transfer data sample sizes, with each figure cell showing the percentage difference in average SMAPE (\%) between the two approaches for specific transfer settings. Positive values indicate better accuracy from transfer learning (i.e. lower SMAPE).
  }
  \label{figure:smape_heatmap_real_world_porkchop_plot_Earth_Mars}
\end{figure*}

\subsection{Transferring RFR on BBOB}
Fig.~\ref{figure:smape_heatmap} shows the average SMAPE differences (\%) between RFR trained from scratch and the transferred RFR. In most cases, the transferred model achieves better fits, as indicated by positive differences. Notably, the performance gap narrows with increasing sample size, indicating that the benefits of transfer learning diminish as more data becomes available, allowing the model trained from scratch to catch up. Compared to the transferred GPR results reported in the prior study~\cite{DBLP:conf/gecco/PanV0B024}, the RFR exhibits more stable performance, with no extreme outliers observed in the 10-dimensional (10-D) case using 50 transfer samples. This demonstrates that the RFR can still deliver reasonable results even when data is sparse in high-dimensional scenarios. However, achieving effective transfer for certain functions remains challenging. For instance, across various dimensions and transfer data sizes, the transferred RFR consistently struggles within function F24, where the model trained from scratch consistently outperforms it. Similarly, functions F21–F23, especially in high-dimensional scenarios, present substantial challenges for the transferred model, limiting its ability to deliver optimal performance.

For a detailed analysis of the 5-D case, TABLE~\ref{table:finalRFR_5D} (see supplementary material~\cite{Anonymous-supp}) reports mean and standard deviation of SMAPE across ten runs for three models: original RFR, transferred RFR, and RFR trained from scratch. Results for 2-D and 10-D are also provided in the supplementary material~\cite{Anonymous-supp}. Two transfer data sizes are evaluated: $\abs{\mathcal{T}}\in\{50, 50d\}$. Statistical significance is assessed using the Kruskal-Wallis test followed by Dunn's posthoc test at the 5\% level. In the table, transferred models outperforming the original are underlined; bold highlights indicate the better model between the transferred model with the model trained from scratch.

In 5-D functions, the transferred RFR generally outperforms both the original and scratch-trained models at transfer sizes of 50 and 250, with rare exceptions. Notably, for F23 with 50 transfer samples, the original model significantly outperforms the transferred one. Analysis shows that the original RFR underfits the source function, making the affine transfer ineffective. These results indicate that transfer learning is most beneficial when the original surrogate already fits the source function reasonably well.

Figure~\ref{figure:RF_SMAPEplot_dim5} (see supplementary material~\cite{Anonymous-supp}) visualizes how model performance in the 5-D case varies with transfer dataset size. For most functions, the transferred RFR outperforms the model trained from scratch up to 250 samples, confirming its effectiveness in low-data regimes. However, exceptions exist: for F24, the model trained from scratch outperforms the transferred one with as few as 40 samples; for F21, it consistently performs better across all sizes. In F16 and F23, the original RFR outperforms both the transferred model and the model trained from scratch across all dataset sizes (10–250 samples).

Next, we examine how function dimensionality affects the performance of our transfer learning approach. In 2-D (see Fig.~\ref{figure:RFRSMAPEplot_dim2} and Table~\ref{table:finalRFR_2D} in the supplementary material~\cite{Anonymous-supp}), the transferred model generally outperforms the model trained from scratch at small sample sizes (e.g., 50), with F23 and F24 as exceptions. This advantage mostly holds with more data, though F19 and F24 also become exceptions. The transferred model consistently outperforms the original model, except on F23. In 10-D (see supplementary material~\cite{Anonymous-supp}), the transferred RFR shows a clear advantage at small sizes (e.g., 50 samples), but the gap narrows as more data (e.g., 500 samples) becomes available. Transfer is less effective on complex, multi-modal functions like F16 and F21–F24, where performance remains limited.

\subsection{Transferring RFR on Real-world applications}

\begin{figure*}[!ht]
  \centering
  \includegraphics[width=\textwidth]{Plots/smape_heatmap_Real_world_robot_arm_21D.pdf}
  \caption{
  The comparison evaluates Random forest regression models obtained through transfer learning against those trained from scratch on the Kinematics of a Robot Arm. It examines different transfer data sample sizes, with each figure cell showing the percentage difference in average SMAPE (\%) between the two approaches for specific transfer settings. Positive values indicate better accuracy from transfer learning (i.e. lower SMAPE).}
  \label{figure:smape_heatmap_real_world_robot_arm_21D}
\end{figure*}

\begin{figure*}[!ht]
  \centering
  \includegraphics[width=\textwidth]{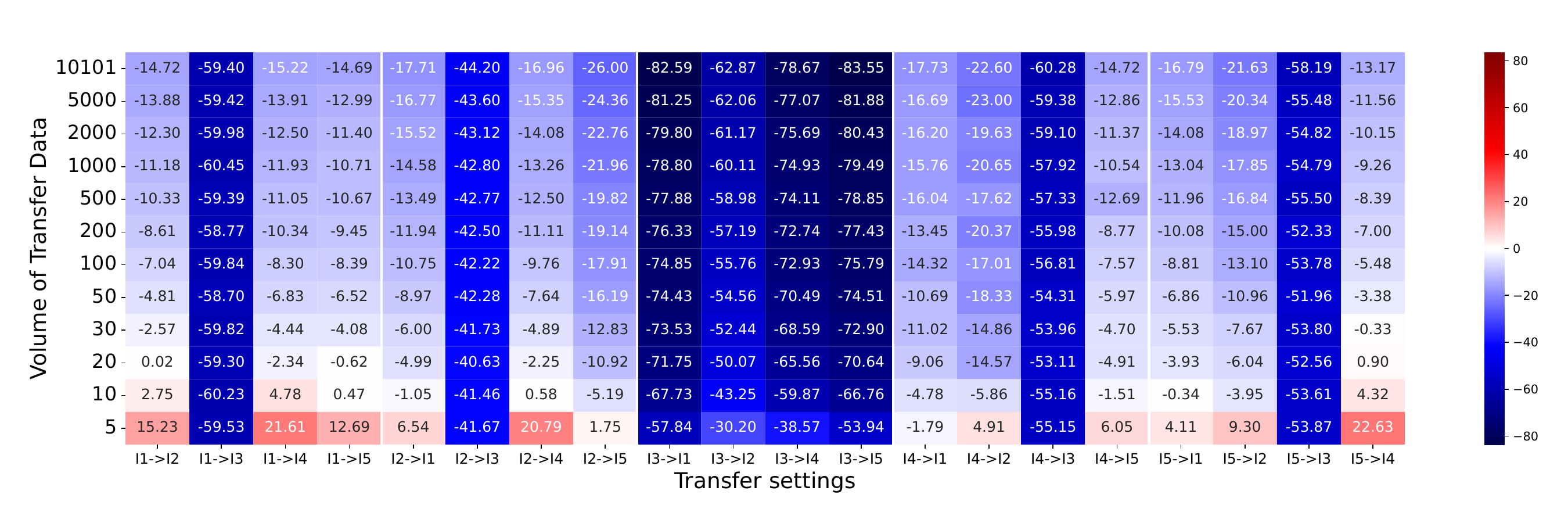}
  \caption{
  The comparison evaluates Random forest regression models obtained through transfer learning against those trained from scratch on the Real-World Optimization Benchmark in Vehicle Dynamics. It examines different transfer data sample sizes, with each figure cell showing the percentage difference in average SMAPE (\%) between the two approaches for specific transfer settings. Positive values indicate better accuracy from transfer learning (i.e. lower SMAPE).}
  \label{figure:smape_heatmap_real_world_vehicle}
\end{figure*}

\subsubsection{Porkchop Plot Benchmarks in Interplanetary Trajectory Optimization}
Fig.~\ref{figure:smape_heatmap_real_world_porkchop_plot_Earth_Mars} shows the SMAPE percentage differences between RFR models trained from scratch and those adapted via transfer learning across various transfer scenarios and dataset sizes. The transferred model typically outperforms the model trained from scratch when fewer than 100 samples are available, highlighting its advantage in data-scarce settings. As the transfer dataset grows, the performance gap narrows, with the models trained from scratch catching up. Fig.~\ref{figure:RFRSMAPEplot_porkchop_plot_Earth_Mars} (see supplementary material~\cite{Anonymous-supp}) offers a detailed view of this trend. In most cases, the transferred model also outperforms the original RFR.

\begin{figure*}[htbp]
  \centering
  \includegraphics[width=\textwidth]{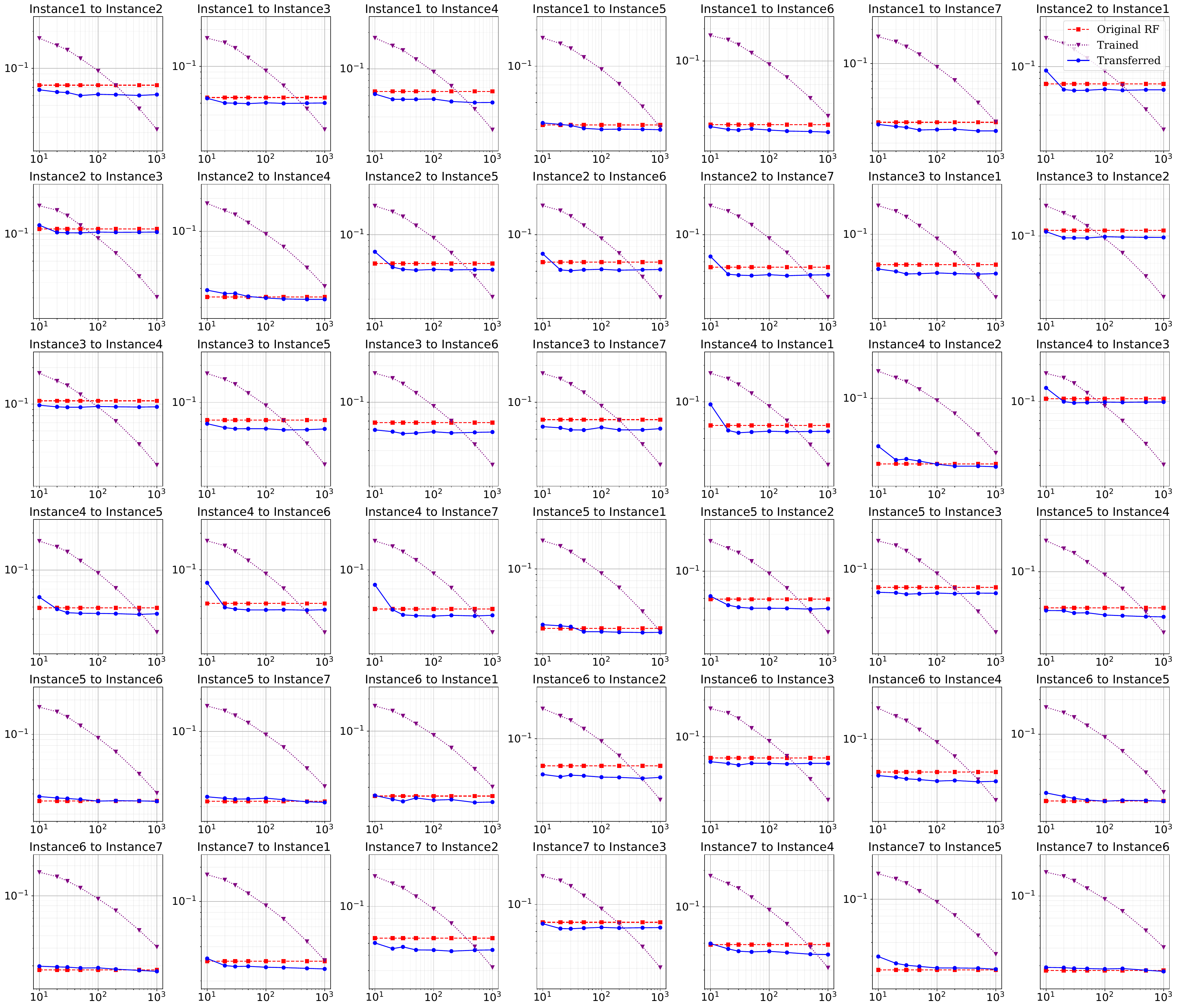}
  \caption{
  The SMAPE values (displayed on the $y$-axis) for three model variants—original RFR, transferred RFR, and a model trained exclusively on the transfer dataset—are analyzed for the Earth-to-Venus mission using Porkchop Plot Benchmarks in Interplanetary Trajectory Optimization. These values are plotted against transfer dataset sizes (shown on the $x$-axis), which include 10, 20, 30, 50, 100, 200, 500, and $1\,000$ samples.
  }
  \label{figure:RFRSMAPEplot_porkchop_plot_earth_venus}
\end{figure*}

The original RFR model rarely surpasses the transferred or scratch-trained models when the transfer dataset is extremely limited, such as with just 10 samples, though occasional exceptions are observed. In such exceptions, the transferred model may overfit the limited data during optimization, leading to poor generalization to the broader target domain. Overfitting can be identified when the transferred model performs better than the original RFR on the transfer samples, as it is explicitly optimized for them. These findings highlight that while the transfer learning approach is generally robust, its effectiveness can be challenged in scenarios with severely restricted data availability. Similar performance patterns are observed in other experimental scenarios, such as Earth-to-Venus and Mercury-to-Earth transfers, as shown in Fig.~\ref{figure:smape_heatmap_real_word_porkchop_plot_earth_venus}, \ref{figure:RFRSMAPEplot_porkchop_plot_earth_venus}, \ref{figure:smape_heatmap_real_word_porkchop_plot_Mercury_Earth}, and \ref{figure:RFRSMAPEplot_porkchop_plot_Mercury_Earth}, which are included in the supplementary material~\cite{Anonymous-supp}.

\subsubsection{Kinematics of a Robot Arm}
As shown in Fig.~\ref{figure:smape_heatmap_real_world_robot_arm_21D}, the results highlight varying outcomes of the proposed transfer learning approach depending on the specific transfer scenario. For example, when transferring from torque1 to torque3, the transferred RFR consistently delivers strong performance regardless of the transfer dataset size. In contrast, scenarios such as transferring from torque7 to torque1 demonstrate a clear advantage for the model trained from scratch, which outperforms the transferred RFR across all dataset sizes. A more granular analysis of how performance correlates with the size of the transfer dataset is provided in Fig.~\ref{figure:RF_SMAPEplot_real_world_kinematics_robot_arm} (see the supplementary material~\cite{Anonymous-supp}). Beyond the general trends, there are rare instances where the original RFR outperforms both the transferred model and the from-scratch model. Such cases are primarily observed when the transfer dataset is extremely small. This behavior mirrors patterns seen in the Interplanetary Trajectory Optimization application, where an insufficient transfer dataset can lead to suboptimal adaptation and even misguide the optimization process.

\subsubsection{Real-World Optimization Benchmark in Vehicle Dynamics}
Compared to other real-world applications, this problem poses significant challenges, as illustrated in Fig.~\ref{figure:smape_heatmap_real_world_vehicle}. In most scenarios, the transferred RFR outperforms the model trained from scratch only when the transfer dataset is relatively small (fewer than 30 samples) and for specific transfer settings. However, there are instances, such as transferring from instance4 to instance3, where the transfer learning approach fails entirely. A more detailed analysis, provided in Fig.~\ref{figure:RF_SMAPEplot_real_world_vehicle_2D} (see the supplementary material~\cite{Anonymous-supp}), shows the SMAPE trends for the original RFR, the model trained from scratch, and the transferred RFR across different transfer dataset sizes. The results underscore the unique difficulty of transferring from instacne3 to other instances. These findings, combined with the landscapes of different instances discussed in the original paper~\cite{DBLP:conf/ijcci/ThomaserVBK23}, suggest that the instance differs substantially from the others. This distinction likely explains why increasing the transfer dataset size—even to encompass the entire domain—fails to meaningfully reduce the SMAPE of the transferred RFR.

\begin{figure*}[!ht]
  \centering
  \includegraphics[width=\textwidth]{Plots/smape_heatmap_Real_world_decoration_frequency_overworld.pdf}
  \caption{
  The comparison evaluates Random forest regression models obtained through transfer learning against those trained from scratch on F5 of the MarioGAN Suite. It examines various transfer data sample sizes, with each figure cell showing the percentage difference in average SMAPE (\%) between the two approaches for specific transfer settings. Positive values indicate better accuracy from transfer learning (i.e. lower SMAPE).}
  \label{figure:smape_heatmap_real_world_decoration_frequency_overworld}
\end{figure*}

\subsubsection{Single-Objective Game-Benchmark MarioGAN Suite}
\begin{figure*}[!ht]
  \centering
  \includegraphics[width=\textwidth]{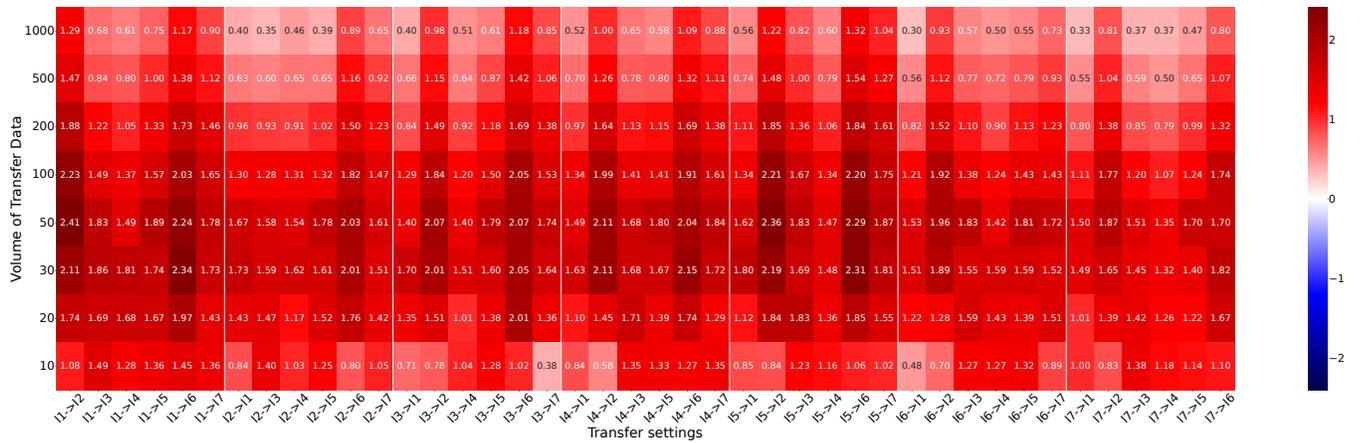}
  \caption{
  The comparison evaluates the performance of Random forest regression (RFR) models obtained through transfer learning against those trained from scratch on F6 of the MarioGAN Suite. This analysis examines various transfer data sample sizes. Each cell in the figure shows the percentage difference in average SMAPE (\%) between the two approaches for specific transfer settings and sample sizes. Positive values indicate that the transferred model achieves superior accuracy by yielding a lower SMAPE compared to the model trained from scratch.}
  \label{figure:smape_heatmap_real_world_decoration_frequency_underground}
\end{figure*}

\begin{figure*}[htbp]
  \centering
  \includegraphics[width=\textwidth]{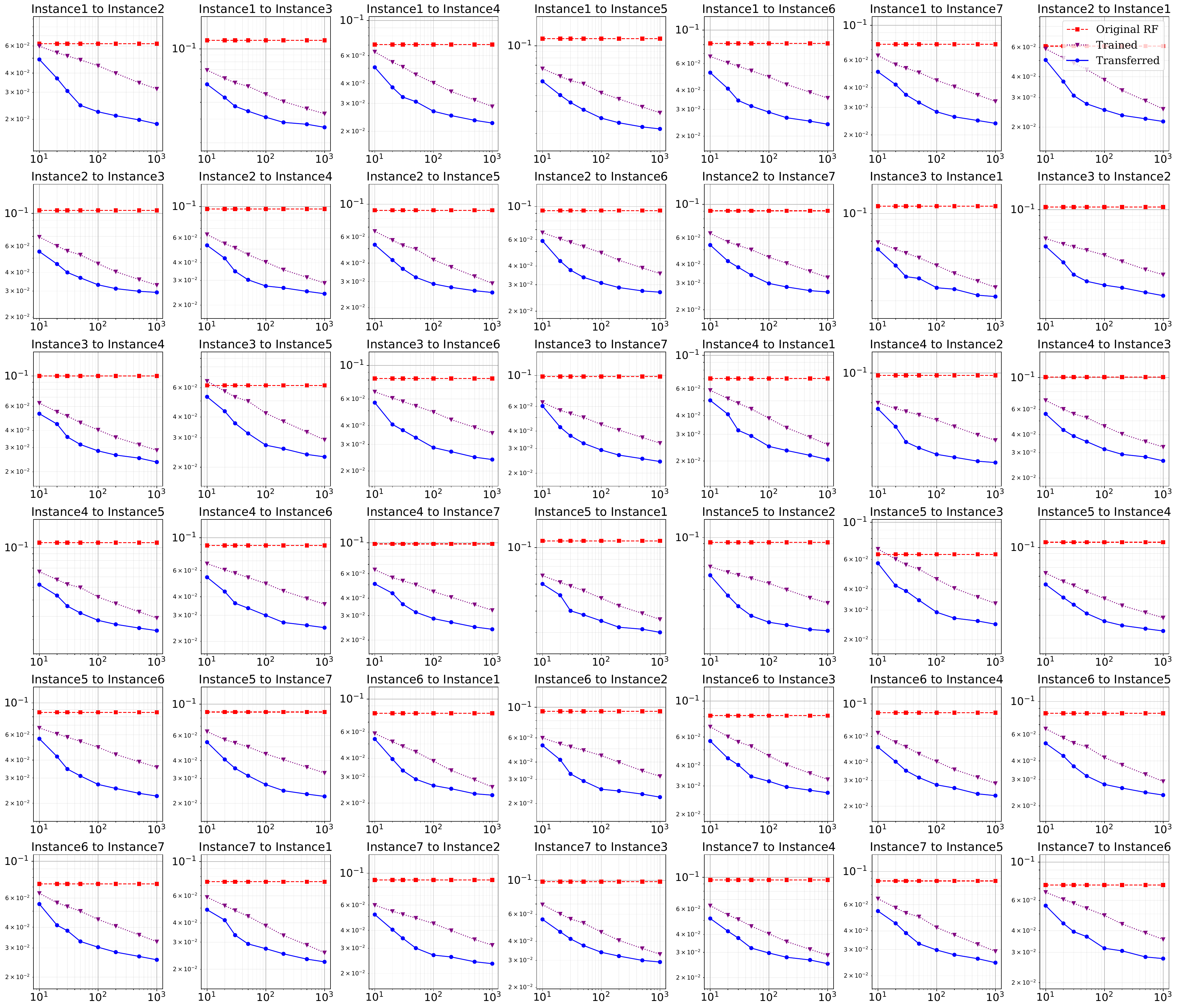}
  \caption{
  The SMAPE values (displayed on the $y$-axis) for three model variants—original RFR, transferred RFR, and a model trained exclusively on the transfer dataset—are analyzed on F6 of Single-Objective Game-Benchmark MarioGAN Suite. These values are plotted against transfer dataset sizes (shown on the $x$-axis), which include 10, 20, 30, 50, 100, 200, 500, and $1\,000$ samples.
  }
  \label{figure:RFRSMAPEplot_Real_world_decoration_frequency_underground}
\end{figure*}

\begin{figure*}[!ht]
  \centering
  \includegraphics[width=\textwidth]{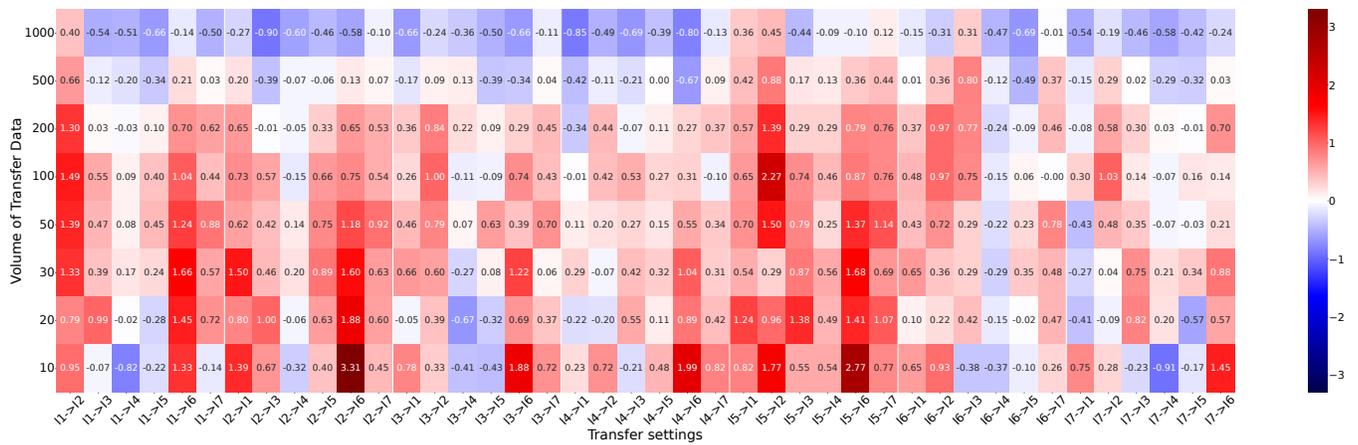}
  \caption{
  The comparison evaluates the performance of Random forest regression (RFR) models obtained through transfer learning against those trained from scratch on F1 of the MarioGAN Suite. This analysis examines various transfer data sample sizes. Each cell in the figure shows the percentage difference in average SMAPE (\%) between the two approaches for specific transfer settings and sample sizes. Positive values indicate that the transferred model achieves superior accuracy by yielding a lower SMAPE compared to the model trained from scratch.}
  \label{figure:smape_heatmap_real_world_enemy_distribution_underground}
\end{figure*}

As illustrated in Fig.~\ref{figure:smape_heatmap_real_world_decoration_frequency_overworld} and Fig.~\ref{figure:RFRSMAPEplot_Real_world_decoration_frequency_overworld} (see the supplementary material~\cite{Anonymous-supp}), the transferred RFR consistently outperforms both the original RFR and the model trained from scratch across most transfer scenarios involving F5 of the MarioGAN Suite. Notably, the performance gap between the transferred RFR and the model trained from scratch becomes more pronounced as the size of the transfer dataset increases up to 30. However, this gap diminishes when the transfer dataset approaches a size of $1\,000$ in the majority of transfer settings. A comparable trend is observed for F1 and F6 of the MarioGAN Suite, as shown in Fig.~\ref{figure:smape_heatmap_real_world_decoration_frequency_underground}, \ref{figure:RFRSMAPEplot_Real_world_decoration_frequency_underground}, \ref{figure:smape_heatmap_real_world_enemy_distribution_underground}, and \ref{figure:RFRSMAPEplot_enemy_distribution_underground}, included in the supplementary material~\cite{Anonymous-supp}.

\section{Conclusion}
We propose a transfer learning method tailored for non-differentiable surrogate models like random forest regression (RFR). This method addresses the challenge of domain shifts between source and target functions connected through an unknown affine transformation. Using a small transfer dataset from the target domain, the method optimizes this transformation to adapt a source-trained surrogate. It offers improved sample efficiency over training from scratch, making it well-suited for data-scarce scenarios. Experiments on BBOB functions show that the transferred RFR outperforms scratch-trained models with just 50–100 transfer samples, especially in high dimensions. However, for complex functions like F16 and F21–F24, transfer learning is less effective due to poor source model accuracy. We further validate the method on four real-world tasks, demonstrating its strength in low-data settings. However, very small transfer sets (e.g., 10 samples) risk overfitting, and large domain shifts limit adaptability even with more data. These findings highlight the need for well-aligned transfer datasets.


Future work could extend this approach to other non-differentiable surrogate models and integrate active learning to jointly refine the surrogate and transformation using informative samples. While our method performs well under affine shifts, its effectiveness drops in cases with complex, nonlinear domain changes (e.g., Vehicle Dynamics). Exploring nonlinear transformations, such as input-output warping~\cite{pmlr-v32-snoek14}, may broaden applicability to more challenging real-world tasks.

\bibliographystyle{IEEEtran}
\bibliography{sample_base}

\clearpage
\begin{table*}[!ht]
\caption{On 2-dimensional BBOB functions, we evaluate the SMAPE metric across three modeling approaches: the original model, the transferred model, and the model trained from scratch using the transfer dataset. The results. reported as the mean and standard deviation over ten repetitions, are provided for two dataset sizes, $\abs{\mathcal{T}}\in\{50, 50d\}$. Statistical comparisons are performed using the Kruskal-Wallis test, followed by Dunn's post-hoc analysis with a significance threshold of 5\%. To emphasize significant differences, we adopt the following conventions: an underlined transferred model indicates it significantly outperforms the original model, while bold font highlights the superior model between the transferred model and the one trained from scratch.
}
\centering
\fontsize{9}{10}\selectfont
\setlength{\tabcolsep}{5pt}
\begin{tabular}{c|ccccc}
    \toprule 
    2D & Original RFR & Train from scratch & Transferred & Train from scratch & Transferred \\
    \cmidrule(lr){3-4} \cmidrule(lr){5-6}
    & & \multicolumn{2}{c}{50 samples} & \multicolumn{2}{c}{100 samples} \\
    \midrule
    F1 & 0.2960 $\pm$ 0.0772 & 0.1219 $\pm$ 0.0239 & \underline{\textbf{0.0212 $\pm$ 0.0015}} & 0.0779 $\pm$ 0.0085 & \underline{\textbf{0.0215 $\pm$ 0.0025}} \\
    F2 & 0.1191 $\pm$ 0.0311 & 0.0516 $\pm$ 0.0167 & \underline{\textbf{0.0050 $\pm$ 0.0035}} & 0.0423 $\pm$ 0.0144 & \underline{\textbf{0.0048 $\pm$ 0.0033}} \\
    F3 & 0.2495 $\pm$ 0.0796 & 0.0654 $\pm$ 0.0117 & \underline{\textbf{0.0295 $\pm$ 0.0058}} & 0.0516 $\pm$ 0.0102 & \underline{\textbf{0.0269 $\pm$ 0.0040}} \\
    F4 & 0.2290 $\pm$ 0.0774 & 0.0597 $\pm$ 0.0123 & \underline{\textbf{0.0262 $\pm$ 0.0071}} & 0.0457 $\pm$ 0.0083 & \underline{0.0278 $\pm$ 0.0106} \\
    F5 & 0.2450 $\pm$ 0.0782 & 0.0409 $\pm$ 0.0152 & \underline{\textbf{0.0052 $\pm$ 0.0024}} & 0.0267 $\pm$ 0.0101 & \underline{\textbf{0.0049 $\pm$ 0.0021}} \\
    F6 & 0.3753 $\pm$ 0.0966 & 0.0923 $\pm$ 0.0202 & \underline{\textbf{0.0221 $\pm$ 0.0036}} & 0.0650 $\pm$ 0.0109 & \underline{\textbf{0.0229 $\pm$ 0.0049}} \\
    F7 & 0.3564 $\pm$ 0.1164 & 0.1830 $\pm$ 0.0304 & \underline{\textbf{0.0633 $\pm$ 0.0074}} & 0.1343 $\pm$ 0.0197 & \underline{\textbf{0.0626 $\pm$ 0.0086}} \\
    F8 & 0.3098 $\pm$ 0.0702 & 0.1489 $\pm$ 0.0103 & \underline{\textbf{0.0422 $\pm$ 0.0083}} & 0.1163 $\pm$ 0.0087 & \underline{\textbf{0.0459 $\pm$ 0.0175}} \\
    F9 & 0.2769 $\pm$ 0.0706 & 0.1401 $\pm$ 0.0136 & \underline{0.0678 $\pm$ 0.0415} & 0.1109 $\pm$ 0.0112 & \underline{0.0730 $\pm$ 0.0531} \\
    F10 & 0.1667 $\pm$ 0.0423 & 0.0575 $\pm$ 0.0182 & \underline{\textbf{0.0153 $\pm$ 0.0020}} & 0.0434 $\pm$ 0.0096 & \underline{\textbf{0.0149 $\pm$ 0.0016}} \\
    F11 & 0.1816 $\pm$ 0.0569 & 0.0451 $\pm$ 0.0089 & \underline{\textbf{0.0153 $\pm$ 0.0038}} & 0.0369 $\pm$ 0.0065 & \underline{\textbf{0.0151 $\pm$ 0.0036}} \\
    F12 & 0.3175 $\pm$ 0.1044 & 0.0551 $\pm$ 0.0112 & \underline{\textbf{0.0049 $\pm$ 0.0023}} & 0.0425 $\pm$ 0.0113 & \underline{\textbf{0.0048 $\pm$ 0.0021}} \\
    F13 & 0.1744 $\pm$ 0.0407 & 0.0581 $\pm$ 0.0171 & \underline{\textbf{0.0154 $\pm$ 0.0022}} & 0.0435 $\pm$ 0.0127 & \underline{\textbf{0.0144 $\pm$ 0.0021}} \\
    F14 & 0.5312 $\pm$ 0.1455 & 0.1643 $\pm$ 0.0392 & \underline{\textbf{0.0365 $\pm$ 0.0125}} & 0.1100 $\pm$ 0.0263 & \underline{\textbf{0.0305 $\pm$ 0.0087}} \\
    F15 & 0.3250 $\pm$ 0.0981 & 0.0762 $\pm$ 0.0102 & \underline{\textbf{0.0339 $\pm$ 0.0049}} & 0.0564 $\pm$ 0.0073 & \underline{\textbf{0.0332 $\pm$ 0.0062}} \\
    F16 & 0.2314 $\pm$ 0.0250 & 0.1706 $\pm$ 0.0144 & \underline{\textbf{0.1026 $\pm$ 0.0127}} & 0.1578 $\pm$ 0.0096 & \underline{\textbf{0.0969 $\pm$ 0.0057}} \\
    F17 & 0.5371 $\pm$ 0.1604 & 0.1460 $\pm$ 0.0164 & \underline{0.1267 $\pm$ 0.0069} & 0.1341 $\pm$ 0.0082 & \underline{0.1185 $\pm$ 0.0080} \\
    F18 & 0.4140 $\pm$ 0.1237 & 0.1072 $\pm$ 0.0086 & \underline{\textbf{0.0864 $\pm$ 0.0046}} & 0.0982 $\pm$ 0.0052 & \underline{0.0849 $\pm$ 0.0032} \\
    F19 & 0.3579 $\pm$ 0.0794 & 0.2229 $\pm$ 0.0128 & \underline{0.2057 $\pm$ 0.0196} & 0.2016 $\pm$ 0.0069 & \underline{0.2025 $\pm$ 0.0142} \\
    F20 & 0.4282 $\pm$ 0.1314 & 0.1533 $\pm$ 0.0266 & \underline{\textbf{0.0345 $\pm$ 0.0031}} & 0.1168 $\pm$ 0.0238 & \underline{\textbf{0.0342 $\pm$ 0.0047}} \\
    F21 & 0.2996 $\pm$ 0.0296 & 0.2402 $\pm$ 0.0177 & \underline{0.1857 $\pm$ 0.0783} & 0.2213 $\pm$ 0.0143 & \underline{0.1818 $\pm$ 0.0743} \\
    F22 & 0.3196 $\pm$ 0.0355 & 0.2069 $\pm$ 0.0246 & \underline{\textbf{0.0961 $\pm$ 0.0125}} & 0.1807 $\pm$ 0.0143 & \underline{\textbf{0.0925 $\pm$ 0.0123}} \\
    F23 & 0.1638 $\pm$ 0.0039 & 0.1717 $\pm$ 0.0095 & 0.1660 $\pm$ 0.0079 & 0.1692 $\pm$ 0.0076 & 0.1655 $\pm$ 0.0043 \\
    F24 & 0.1922 $\pm$ 0.0404 & 0.1383 $\pm$ 0.0242 & 0.1568 $\pm$ 0.0336 & 0.1257 $\pm$ 0.0219 & 0.1583 $\pm$ 0.0371 \\
    \bottomrule
\end{tabular}%
\label{table:finalRFR_2D}
\end{table*}

\begin{table*}[!ht]
\caption{On 5-dimensional BBOB functions, we evaluate the SMAPE metric across three modeling approaches: the original model, the transferred model, and the model trained from scratch using the transfer dataset. The results. reported as the mean and standard deviation over ten repetitions, are provided for two dataset sizes, $\abs{\mathcal{T}}\in\{50, 50d\}$. Statistical comparisons are performed using the Kruskal-Wallis test, followed by Dunn's post-hoc analysis with a significance threshold of 5\%. To emphasize significant differences, we adopt the following conventions: an underlined transferred model indicates it significantly outperforms the original model, while bold font highlights the superior model between the transferred model and the one trained from scratch.
}
\centering
\fontsize{9}{10}\selectfont
\setlength{\tabcolsep}{5pt}
\begin{tabular}{c|ccccc}
    \toprule 
    5D & Original RFR & Train from scratch & Transferred & Train from scratch & Transferred \\
    \cmidrule(lr){3-4} \cmidrule(lr){5-6}
    & & \multicolumn{2}{c}{50 samples} & \multicolumn{2}{c}{250 samples} \\
    \midrule
    F1 & 0.1141 $\pm$ 0.0346 & 0.0748 $\pm$ 0.0050 & \underline{\textbf{0.0350 $\pm$ 0.0056}} & 0.0491 $\pm$ 0.0020 & \underline{\textbf{0.0293 $\pm$ 0.0030}} \\
    F2 & 0.1047 $\pm$ 0.0132 & 0.0611 $\pm$ 0.0094 & \underline{\textbf{0.0093 $\pm$ 0.0053}} & 0.0437 $\pm$ 0.0068 & \underline{\textbf{0.0050 $\pm$ 0.0007}} \\
    F3 & 0.0905 $\pm$ 0.0180 & 0.0554 $\pm$ 0.0048 & \underline{0.0403 $\pm$ 0.0077} & 0.0376 $\pm$ 0.0026 & \underline{\textbf{0.0305 $\pm$ 0.0029}} \\
    F4 & 0.1404 $\pm$ 0.0126 & 0.0960 $\pm$ 0.0082 & \underline{0.0716 $\pm$ 0.0144} & 0.0706 $\pm$ 0.0079 & \underline{\textbf{0.0464 $\pm$ 0.0115}} \\
    F5 & 0.0774 $\pm$ 0.0169 & 0.0261 $\pm$ 0.0043 & \underline{\textbf{0.0088 $\pm$ 0.0008}} & 0.0146 $\pm$ 0.0030 & \underline{\textbf{0.0069 $\pm$ 0.0008}} \\
    F6 & 0.1334 $\pm$ 0.0320 & 0.0844 $\pm$ 0.0115 & \underline{\textbf{0.0552 $\pm$ 0.0100}} & 0.0577 $\pm$ 0.0052 & \underline{0.0470 $\pm$ 0.0075} \\
    F7 & 0.1638 $\pm$ 0.0128 & 0.1179 $\pm$ 0.0130 & \underline{\textbf{0.0691 $\pm$ 0.0067}} & 0.0808 $\pm$ 0.0102 & \underline{0.0572 $\pm$ 0.0034} \\
    F8 & 0.1042 $\pm$ 0.0102 & 0.0726 $\pm$ 0.0041 & \underline{0.0557 $\pm$ 0.0099} & 0.0519 $\pm$ 0.0051 & \underline{0.0397 $\pm$ 0.0039} \\
    F9 & 0.0787 $\pm$ 0.0115 & 0.0748 $\pm$ 0.0048 & \underline{\textbf{0.0524 $\pm$ 0.0053}} & 0.0549 $\pm$ 0.0039 & \underline{\textbf{0.0435 $\pm$ 0.0044}} \\
    F10 & 0.1139 $\pm$ 0.0168 & 0.0576 $\pm$ 0.0067 & \underline{\textbf{0.0315 $\pm$ 0.0033}} & 0.0410 $\pm$ 0.0062 & \underline{\textbf{0.0293 $\pm$ 0.0027}} \\
    F11 & 0.1496 $\pm$ 0.0279 & 0.0926 $\pm$ 0.0104 & \underline{\textbf{0.0421 $\pm$ 0.0053}} & 0.0725 $\pm$ 0.0070 & \underline{\textbf{0.0389 $\pm$ 0.0036}} \\
    F12 & 0.0692 $\pm$ 0.0161 & 0.0336 $\pm$ 0.0026 & \underline{\textbf{0.0256 $\pm$ 0.0029}} & 0.0219 $\pm$ 0.0019 & \underline{0.0208 $\pm$ 0.0024} \\
    F13 & 0.0576 $\pm$ 0.0094 & 0.0257 $\pm$ 0.0047 & \underline{\textbf{0.0169 $\pm$ 0.0018}} & 0.0170 $\pm$ 0.0024 & \underline{0.0124 $\pm$ 0.0021} \\
    F14 & 0.3330 $\pm$ 0.0926 & 0.1371 $\pm$ 0.0118 & \underline{\textbf{0.0915 $\pm$ 0.0110}} & 0.0900 $\pm$ 0.0106 & \underline{0.0740 $\pm$ 0.0059} \\
    F15 & 0.1421 $\pm$ 0.0300 & 0.0572 $\pm$ 0.0047 & \underline{\textbf{0.0338 $\pm$ 0.0038}} & 0.0358 $\pm$ 0.0039 & \underline{\textbf{0.0277 $\pm$ 0.0030}} \\
    F16 & 0.1066 $\pm$ 0.0012 & 0.1139 $\pm$ 0.0043 & \textbf{0.1087 $\pm$ 0.0026} & 0.1096 $\pm$ 0.0025 & 0.1080 $\pm$ 0.0021 \\
    F17 & 0.3131 $\pm$ 0.0405 & 0.1322 $\pm$ 0.0176 & \underline{0.1024 $\pm$ 0.0093} & 0.0993 $\pm$ 0.0078 & \underline{0.0960 $\pm$ 0.0087} \\
    F18 & 0.2122 $\pm$ 0.0265 & 0.0926 $\pm$ 0.0157 & \underline{0.0715 $\pm$ 0.0050} & 0.0699 $\pm$ 0.0044 & \underline{0.0649 $\pm$ 0.0030} \\
    F19 & 0.1468 $\pm$ 0.0139 & 0.1350 $\pm$ 0.0069 & \underline{\textbf{0.1027 $\pm$ 0.0108}} & 0.0986 $\pm$ 0.0034 & \underline{\textbf{0.0846 $\pm$ 0.0048}} \\
    F20 & 0.1125 $\pm$ 0.0193 & 0.0710 $\pm$ 0.0122 & \underline{\textbf{0.0484 $\pm$ 0.0056}} & 0.0462 $\pm$ 0.0039 & \underline{\textbf{0.0356 $\pm$ 0.0018}} \\
    F21 & 0.0707 $\pm$ 0.0036 & 0.0650 $\pm$ 0.0054 & 0.0702 $\pm$ 0.0027 & \textbf{0.0575 $\pm$ 0.0025} & 0.0664 $\pm$ 0.0026 \\
    F22 & 0.0403 $\pm$ 0.0020 & 0.0386 $\pm$ 0.0070 & 0.0382 $\pm$ 0.0047 & 0.0314 $\pm$ 0.0023 & \underline{0.0337 $\pm$ 0.0019} \\
    F23 & \underline{0.1420 $\pm$ 0.0020} & 0.1501 $\pm$ 0.0075 & 0.1468 $\pm$ 0.0038 & 0.1444 $\pm$ 0.0026 & 0.1438 $\pm$ 0.0031 \\
    F24 & 0.2046 $\pm$ 0.0217 & 0.1987 $\pm$ 0.0183 & 0.2035 $\pm$ 0.0197 & \textbf{0.1698 $\pm$ 0.0145} & 0.1980 $\pm$ 0.0198 \\
    \bottomrule
\end{tabular}%
\label{table:finalRFR_5D}
\end{table*}

\begin{table*}[!ht]
\caption{On 10-dimensional BBOB functions, we evaluate the SMAPE metric across three modeling approaches: the original model, the transferred model, and the model trained from scratch using the transfer dataset. The results. reported as the mean and standard deviation over ten repetitions, are provided for two dataset sizes, $\abs{\mathcal{T}}\in\{50, 50d\}$. Statistical comparisons are performed using the Kruskal-Wallis test, followed by Dunn's post-hoc analysis with a significance threshold of 5\%. To emphasize significant differences, we adopt the following conventions: an underlined transferred model indicates it significantly outperforms the original model, while bold font highlights the superior model between the transferred model and the one trained from scratch.
}
\centering
\fontsize{9}{10}\selectfont
\setlength{\tabcolsep}{5pt}
\begin{tabular}{c|ccccc}
    \toprule
    10D & Original RFR & Train from scratch & Transferred & Train from scratch & Transferred \\
    \cmidrule(lr){3-4} \cmidrule(lr){5-6}
    & & \multicolumn{2}{c}{50 samples} & \multicolumn{2}{c}{500 samples} \\
    \midrule
    F1 & 0.0713 $\pm$ 0.0107 & 0.0509 $\pm$ 0.0037 & \underline{\textbf{0.0381 $\pm$ 0.0020}} & 0.0348 $\pm$ 0.0011 & \underline{0.0325 $\pm$ 0.0011} \\
    F2 & 0.0811 $\pm$ 0.0132 & 0.0465 $\pm$ 0.0041 & \underline{\textbf{0.0220 $\pm$ 0.0064}} & 0.0332 $\pm$ 0.0025 & \underline{\textbf{0.0097 $\pm$ 0.0017}} \\
    F3 & 0.0647 $\pm$ 0.0023 & 0.0475 $\pm$ 0.0036 & \underline{0.0403 $\pm$ 0.0039} & 0.0343 $\pm$ 0.0019 & \underline{\textbf{0.0283 $\pm$ 0.0020}} \\
    F4 & 0.1220 $\pm$ 0.0049 & 0.0861 $\pm$ 0.0035 & \underline{0.0925 $\pm$ 0.0065} & 0.0669 $\pm$ 0.0021 & \underline{\textbf{0.0546 $\pm$ 0.0048}} \\
    F5 & 0.0462 $\pm$ 0.0079 & 0.0204 $\pm$ 0.0027 & \underline{\textbf{0.0094 $\pm$ 0.0003}} & 0.0122 $\pm$ 0.0019 & \underline{\textbf{0.0079 $\pm$ 0.0003}} \\
    F6 & 0.0606 $\pm$ 0.0058 & 0.0334 $\pm$ 0.0033 & \underline{\textbf{0.0230 $\pm$ 0.0009}} & 0.0232 $\pm$ 0.0023 & \underline{0.0194 $\pm$ 0.0011} \\
    F7 & 0.1001 $\pm$ 0.0108 & 0.0575 $\pm$ 0.0021 & \underline{\textbf{0.0459 $\pm$ 0.0028}} & 0.0418 $\pm$ 0.0021 & \underline{\textbf{0.0345 $\pm$ 0.0015}} \\
    F8 & 0.0659 $\pm$ 0.0072 & 0.0468 $\pm$ 0.0028 & \underline{0.0423 $\pm$ 0.0023} & 0.0342 $\pm$ 0.0007 & \underline{0.0330 $\pm$ 0.0015} \\
    F9 & \underline{0.0409 $\pm$ 0.0009} & 0.0481 $\pm$ 0.0021 & 0.0454 $\pm$ 0.0023 & 0.0381 $\pm$ 0.0006 & \underline{0.0381 $\pm$ 0.0017} \\
    F10 & 0.0708 $\pm$ 0.0141 & 0.0387 $\pm$ 0.0030 & \underline{\textbf{0.0250 $\pm$ 0.0025}} & 0.0273 $\pm$ 0.0022 & \underline{\textbf{0.0201 $\pm$ 0.0015}} \\
    F11 & 0.1375 $\pm$ 0.0110 & 0.1273 $\pm$ 0.0121 & \underline{\textbf{0.0763 $\pm$ 0.0037}} & 0.0994 $\pm$ 0.0086 & \underline{\textbf{0.0695 $\pm$ 0.0021}} \\
    F12 & 0.0617 $\pm$ 0.0054 & 0.0370 $\pm$ 0.0011 & \underline{\textbf{0.0304 $\pm$ 0.0013}} & 0.0272 $\pm$ 0.0009 & \underline{0.0254 $\pm$ 0.0012} \\
    F13 & 0.0342 $\pm$ 0.0049 & 0.0179 $\pm$ 0.0016 & \underline{\textbf{0.0143 $\pm$ 0.0009}} & 0.0123 $\pm$ 0.0004 & \underline{\textbf{0.0107 $\pm$ 0.0006}} \\
    F14 & 0.1816 $\pm$ 0.0248 & 0.1052 $\pm$ 0.0073 & \underline{\textbf{0.0832 $\pm$ 0.0052}} & 0.0748 $\pm$ 0.0055 & \underline{\textbf{0.0584 $\pm$ 0.0031}} \\
    F15 & 0.0796 $\pm$ 0.0082 & 0.0496 $\pm$ 0.0037 & \underline{0.0446 $\pm$ 0.0030} & 0.0350 $\pm$ 0.0008 & \underline{0.0338 $\pm$ 0.0013} \\
    F16 & \underline{0.0715 $\pm$ 0.0006} & 0.0772 $\pm$ 0.0037 & 0.0745 $\pm$ 0.0026 & 0.0726 $\pm$ 0.0006 & 0.0728 $\pm$ 0.0009 \\
    F17 & 0.2179 $\pm$ 0.0257 & 0.1238 $\pm$ 0.0082 & \underline{0.1072 $\pm$ 0.0107} & 0.0930 $\pm$ 0.0037 & \underline{0.0839 $\pm$ 0.0053} \\
    F18 & 0.1760 $\pm$ 0.0223 & 0.0949 $\pm$ 0.0060 & \underline{0.0779 $\pm$ 0.0073} & 0.0700 $\pm$ 0.0030 & \underline{\textbf{0.0580 $\pm$ 0.0032}} \\
    F19 & 0.0911 $\pm$ 0.0040 & 0.1026 $\pm$ 0.0033 & 0.0980 $\pm$ 0.0056 & 0.0847 $\pm$ 0.0021 & \underline{0.0826 $\pm$ 0.0025} \\
    F20 & 0.0722 $\pm$ 0.0085 & 0.0429 $\pm$ 0.0015 & \underline{0.0389 $\pm$ 0.0037} & 0.0304 $\pm$ 0.0014 & \underline{0.0270 $\pm$ 0.0023} \\
    F21 & 0.0156 $\pm$ 0.0003 & \textbf{0.0147 $\pm$ 0.0010} & 0.0161 $\pm$ 0.0005 & \textbf{0.0132 $\pm$ 0.0006} & 0.0155 $\pm$ 0.0004 \\
    F22 & 0.0118 $\pm$ 0.0005 & 0.0098 $\pm$ 0.0007 & \underline{0.0107 $\pm$ 0.0004} & \textbf{0.0082 $\pm$ 0.0004} & \underline{0.0101 $\pm$ 0.0004} \\
    F23 & \underline{0.1253 $\pm$ 0.0012} & 0.1290 $\pm$ 0.0052 & 0.1296 $\pm$ 0.0025 & \textbf{0.1214 $\pm$ 0.0013} & 0.1277 $\pm$ 0.0017 \\
    F24 & 0.2761 $\pm$ 0.0058 & \textbf{0.2199 $\pm$ 0.0051} & 0.2732 $\pm$ 0.0056 & \textbf{0.1951 $\pm$ 0.0014} & 0.2713 $\pm$ 0.0051 \\
    \bottomrule
\end{tabular}%
\label{table:finalRFR_2D10D}
\end{table*}

\begin{figure*}[htbp]
  \centering
  \includegraphics[width=\textwidth]{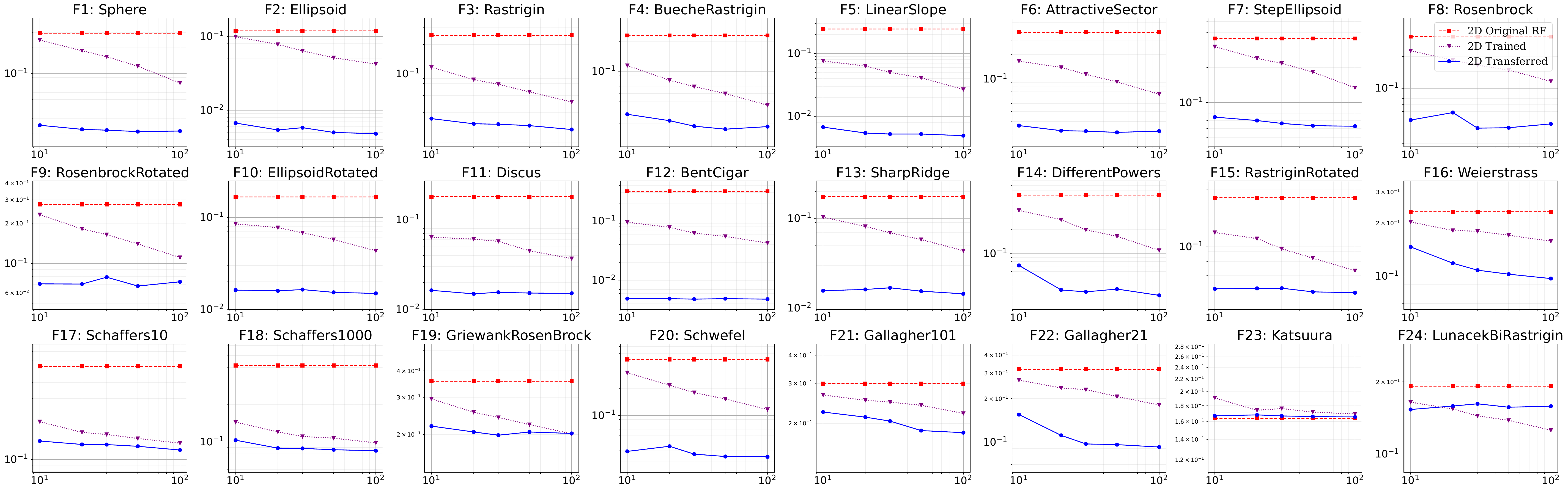}
  \caption{The SMAPE values ($y$-axis) for three model variants --- original RFR, transferred RFR, and the model trained exclusively on the transfer dataset --- are plotted against the size of the transfer dataset on 2-dimensional BBOB functions. The dataset sizes ($x$-axis) considered are 10, 20, 30, 50, and 100 samples.}
  \label{figure:RFRSMAPEplot_dim2}
\end{figure*}

\begin{figure*}[!ht]
  \centering
  \includegraphics[width=\textwidth]{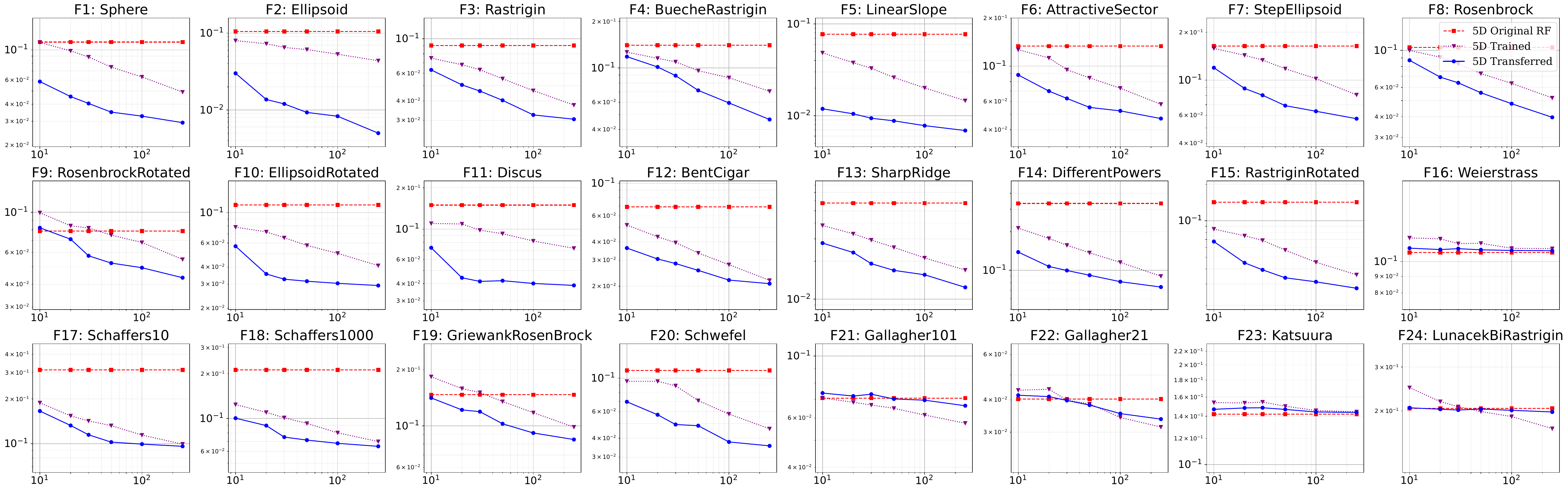}
  \caption{The SMAPE values ($y$-axis) for three model variants --- original RFR, transferred RFR, and the model trained exclusively on the transfer dataset --- are plotted against the size of the transfer dataset on 5-dimensional BBOB functions. The dataset sizes ($x$-axis) considered are 10, 20, 30, 50, 100, and 250 samples.}
  \label{figure:RF_SMAPEplot_dim5}
\end{figure*}

\begin{figure*}[htbp]
  \centering
  \includegraphics[width=\textwidth]{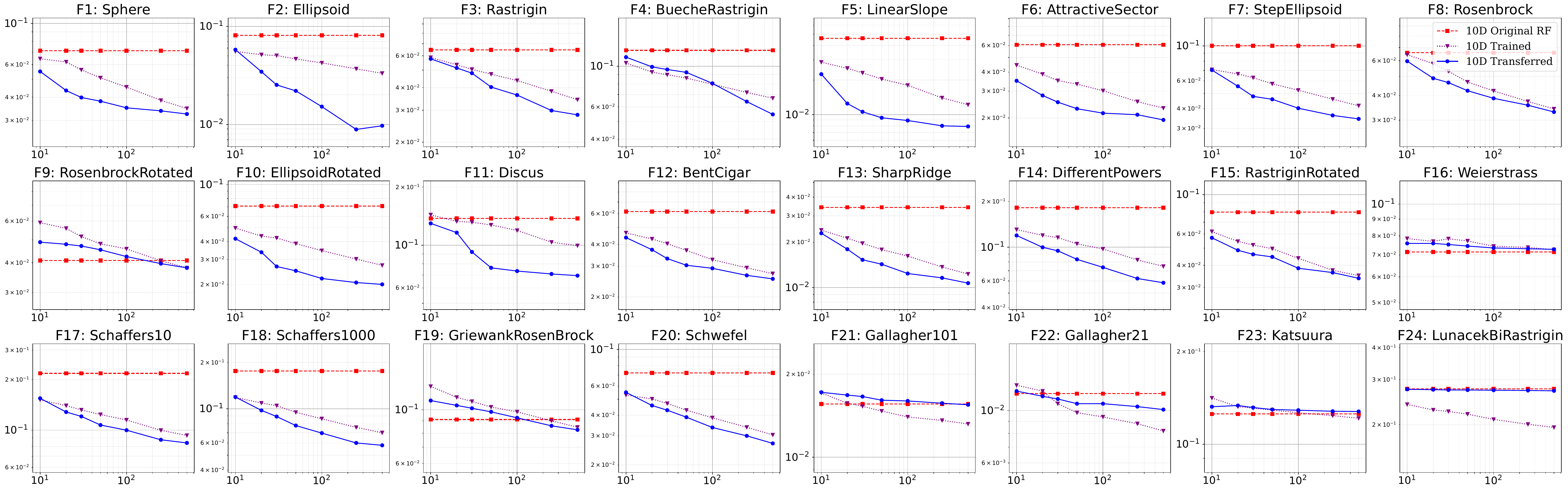}
  \caption{The SMAPE values ($y$-axis) for three model variants --- original RFR, transferred RFR, and the model trained exclusively on the transfer dataset --- are plotted against the size of the transfer dataset on 10-dimensional BBOB functions. The dataset sizes ($x$-axis) considered are 10, 20, 30, 50, 100, 250, and 500 samples.
  }
  \label{figure:RFRSMAPEplot_dim10}
\end{figure*}

\begin{figure*}[htbp]
  \centering
  \includegraphics[width=\textwidth]{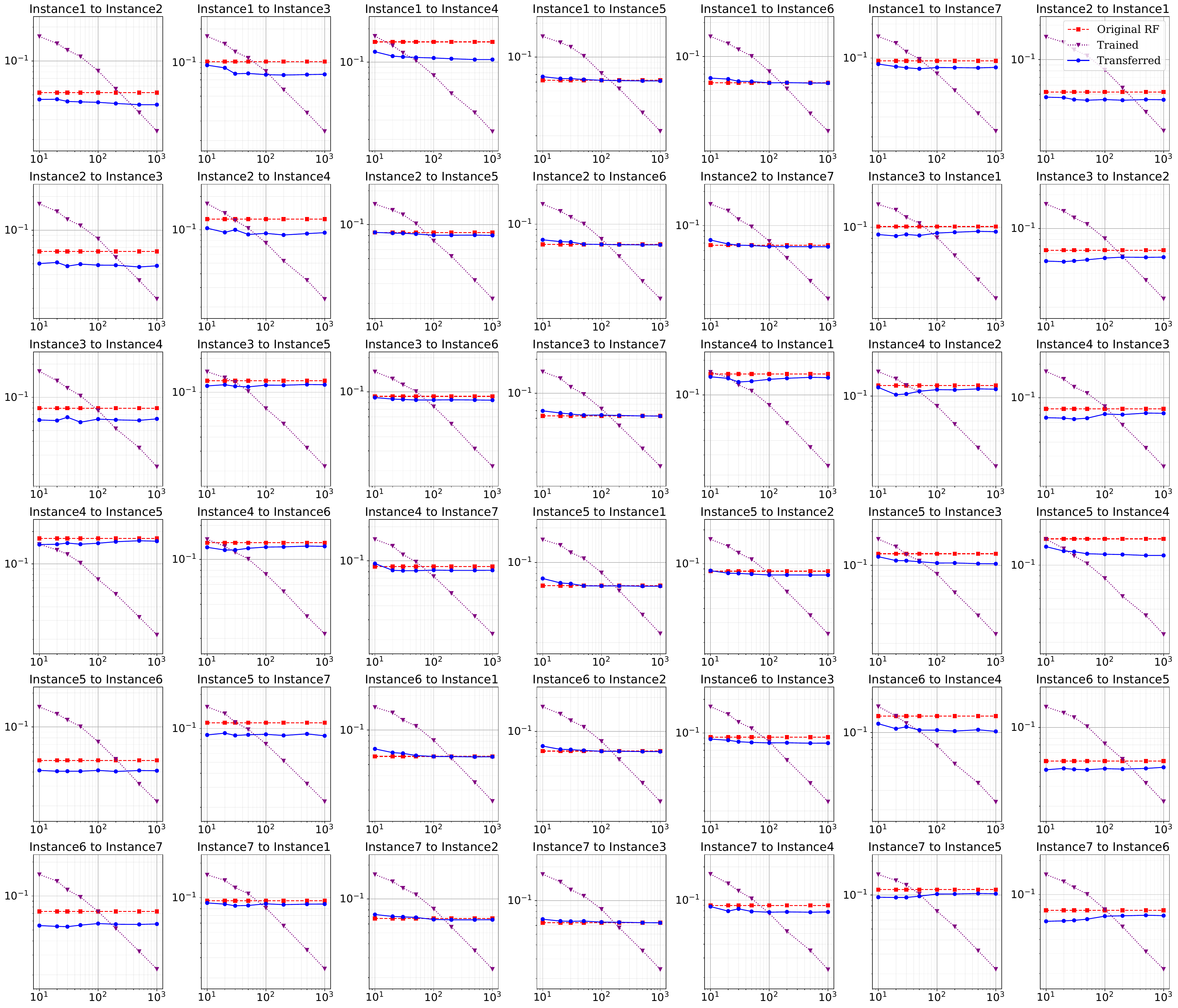}
  \caption{
  The SMAPE values (displayed on the $y$-axis) for three model variants—original RFR, transferred RFR, and a model trained exclusively on the transfer dataset—are analyzed for the Earth-to-Mars mission using Porkchop Plot Benchmarks in Interplanetary Trajectory Optimization. These values are plotted against transfer dataset sizes (shown on the $x$-axis), which include 10, 20, 30, 50, 100, 200, 500, and 1\,000 samples.
  }
  \label{figure:RFRSMAPEplot_porkchop_plot_Earth_Mars}
\end{figure*}
\begin{figure*}[!ht]
  \centering
  \includegraphics[width=\textwidth]{Plots/smape_heatmap_Real_world_Porkchop_plot_Earth_Venus.pdf}
  \caption{
  The comparison evaluates the performance of Random forest regression (RFR) models obtained through transfer learning against those trained from scratch on the Porkchop Plot Benchmarks in Interplanetary Trajectory Optimization, focusing on the Earth-to-Venus mission. This analysis examines various transfer data sample sizes. Each cell in the figure shows the percentage difference in average SMAPE (\%) between the two approaches for specific transfer settings and sample sizes. Positive values indicate that the transferred model achieves superior accuracy by yielding a lower SMAPE compared to the model trained from scratch.}
  \label{figure:smape_heatmap_real_word_porkchop_plot_earth_venus}
\end{figure*}
\begin{figure*}[!ht]
  \centering
  \includegraphics[width=\textwidth]{Plots/smape_heatmap_Real_world_Porkchop_plot_Mercury_Earth.pdf}
  \caption{
  The comparison evaluates the performance of Random forest regression (RFR) models obtained through transfer learning against those trained from scratch on the Porkchop Plot Benchmarks in Interplanetary Trajectory Optimization, focusing on the Mercury-to-Earth mission. This analysis examines various transfer data sample sizes. Each cell in the figure shows the percentage difference in average SMAPE (\%) between the two approaches for specific transfer settings and sample sizes. Positive values indicate that the transferred model achieves superior accuracy by yielding a lower SMAPE compared to the model trained from scratch.}
  \label{figure:smape_heatmap_real_word_porkchop_plot_Mercury_Earth}
\end{figure*}

\begin{figure*}[htbp]
  \centering
  \includegraphics[width=\textwidth]{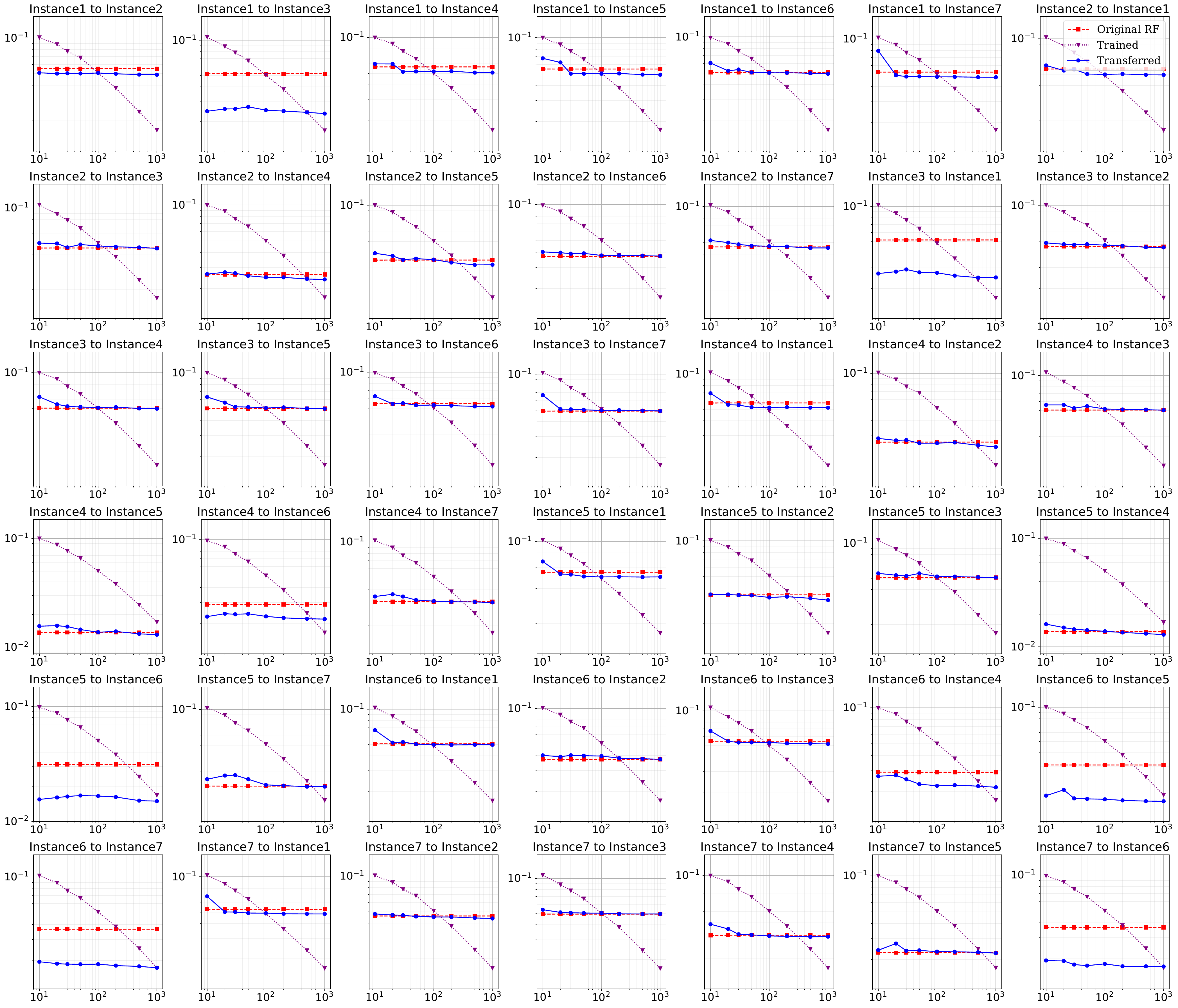}
  \caption{
  The SMAPE values (displayed on the $y$-axis) for three model variants—original RFR, transferred RFR, and a model trained exclusively on the transfer dataset—are analyzed for the Mercury-to-Earth mission using Porkchop Plot Benchmarks in Interplanetary Trajectory Optimization. These values are plotted against transfer dataset sizes (shown on the $x$-axis), which include 10, 20, 30, 50, 100, 200, 500, and $1\,000$ samples.
  }
  \label{figure:RFRSMAPEplot_porkchop_plot_Mercury_Earth}
\end{figure*}

\begin{figure*}[!ht]
  \centering
  \includegraphics[width=\textwidth]{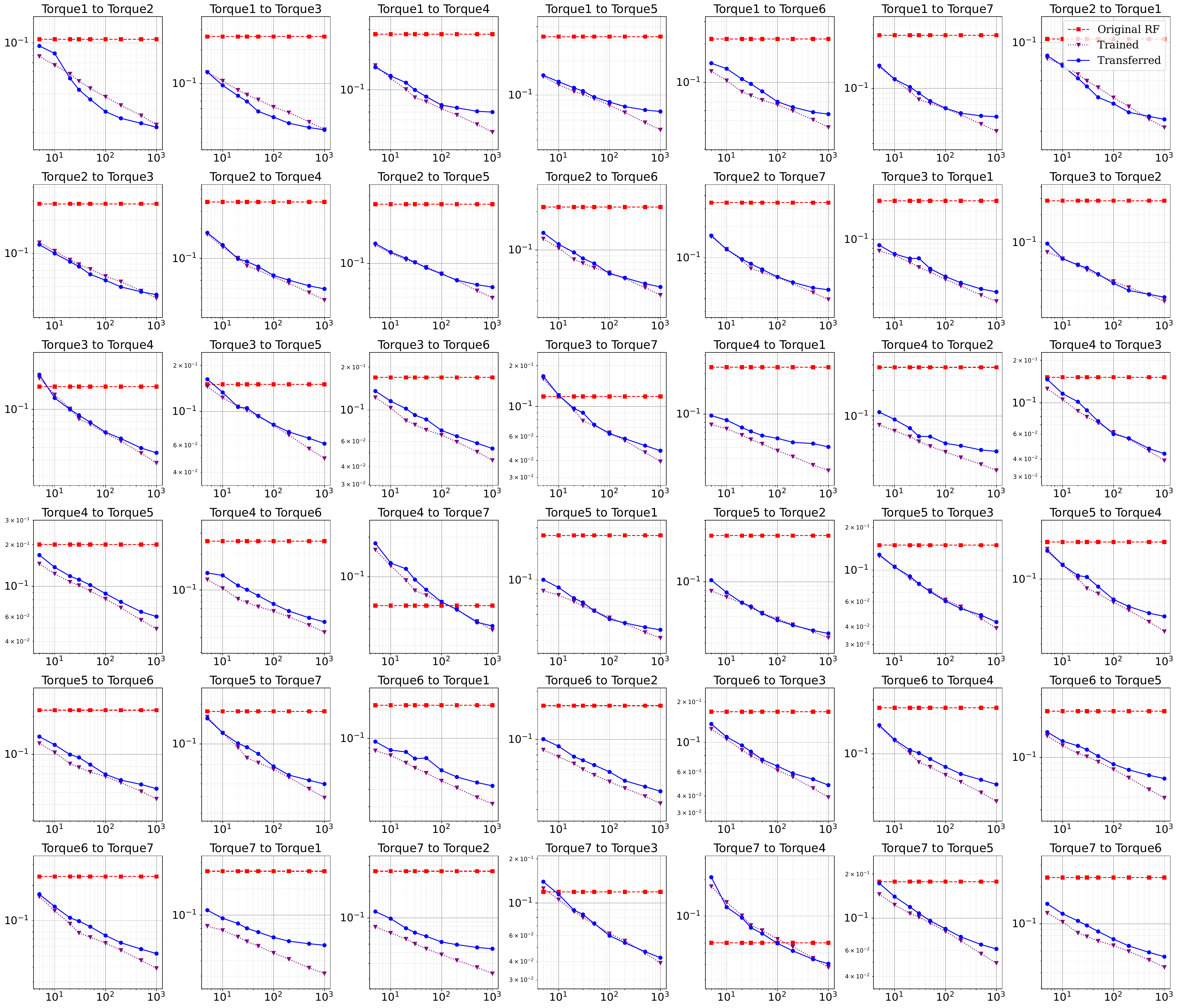}
  \caption{
  The SMAPE values (displayed on the $y$-axis) for three model variants—original RFR, transferred RFR, and a model trained exclusively on the transfer dataset—are analyzed on the Kinematics of the Robot Arm real-world application. These values are plotted against transfer dataset sizes (shown on the $x$-axis), which include 5, 10, 20, 30, 50, 100, 200, 500, and $1\,000$ samples.
  }
  \label{figure:RF_SMAPEplot_real_world_kinematics_robot_arm}
\end{figure*}

\begin{figure*}[!ht]
  \centering
  \includegraphics[width=\textwidth]{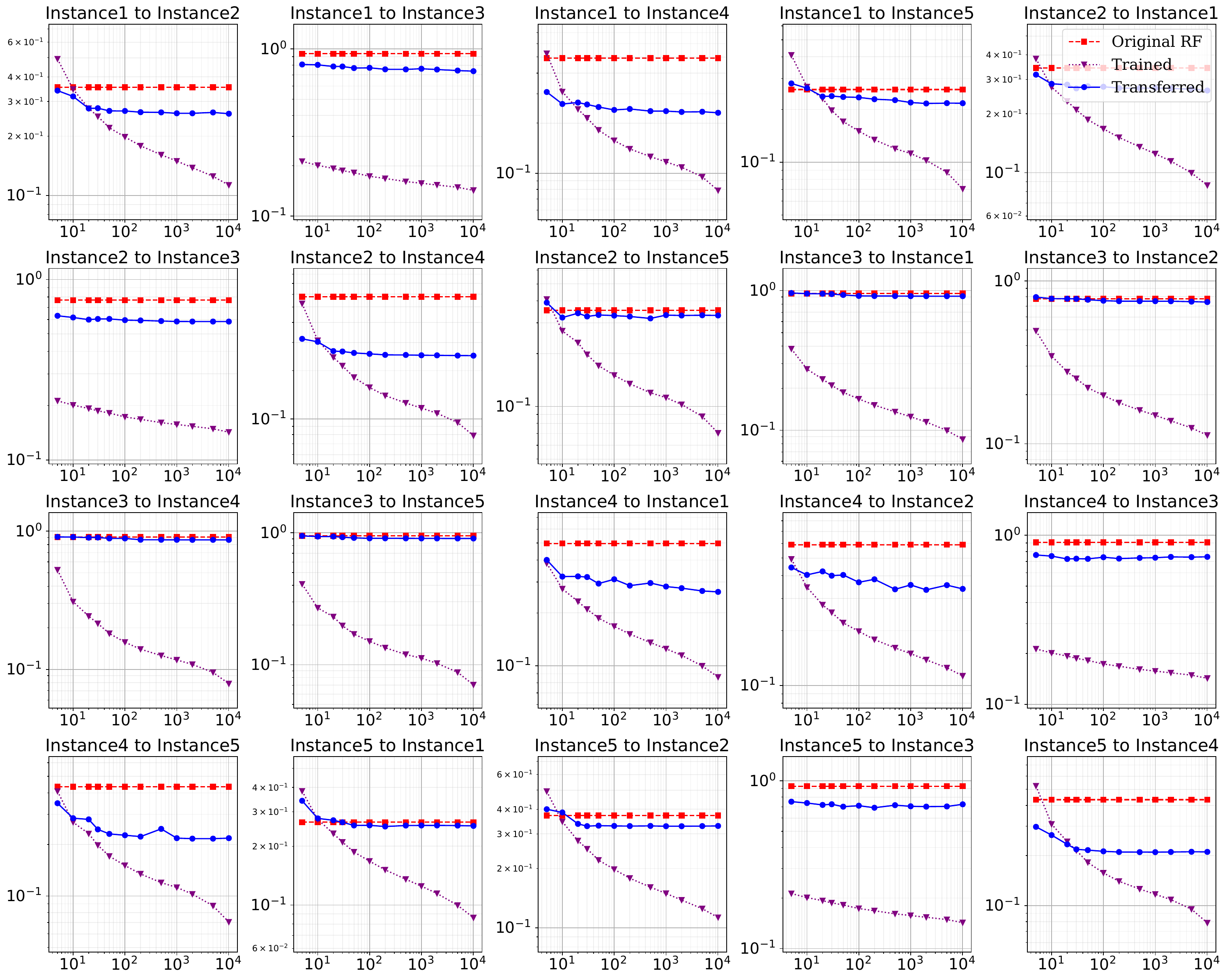}
  \caption{The SMAPE values (displayed on the $y$-axis) for three model variants—original RFR, transferred RFR, and a model trained exclusively on the transfer dataset—are analyzed on the real-world optimization benchmark from vehicle dynamics. These values are plotted against transfer dataset sizes (shown on the $x$-axis), which include 5, 10, 20, 30, 50, 100, 200, 500, 1\,000, 2\,000, 5\,000, and 10\,101 samples. 
  }
  \label{figure:RF_SMAPEplot_real_world_vehicle_2D}
\end{figure*}

\begin{figure*}[htbp]
  \centering
  \includegraphics[width=\textwidth]{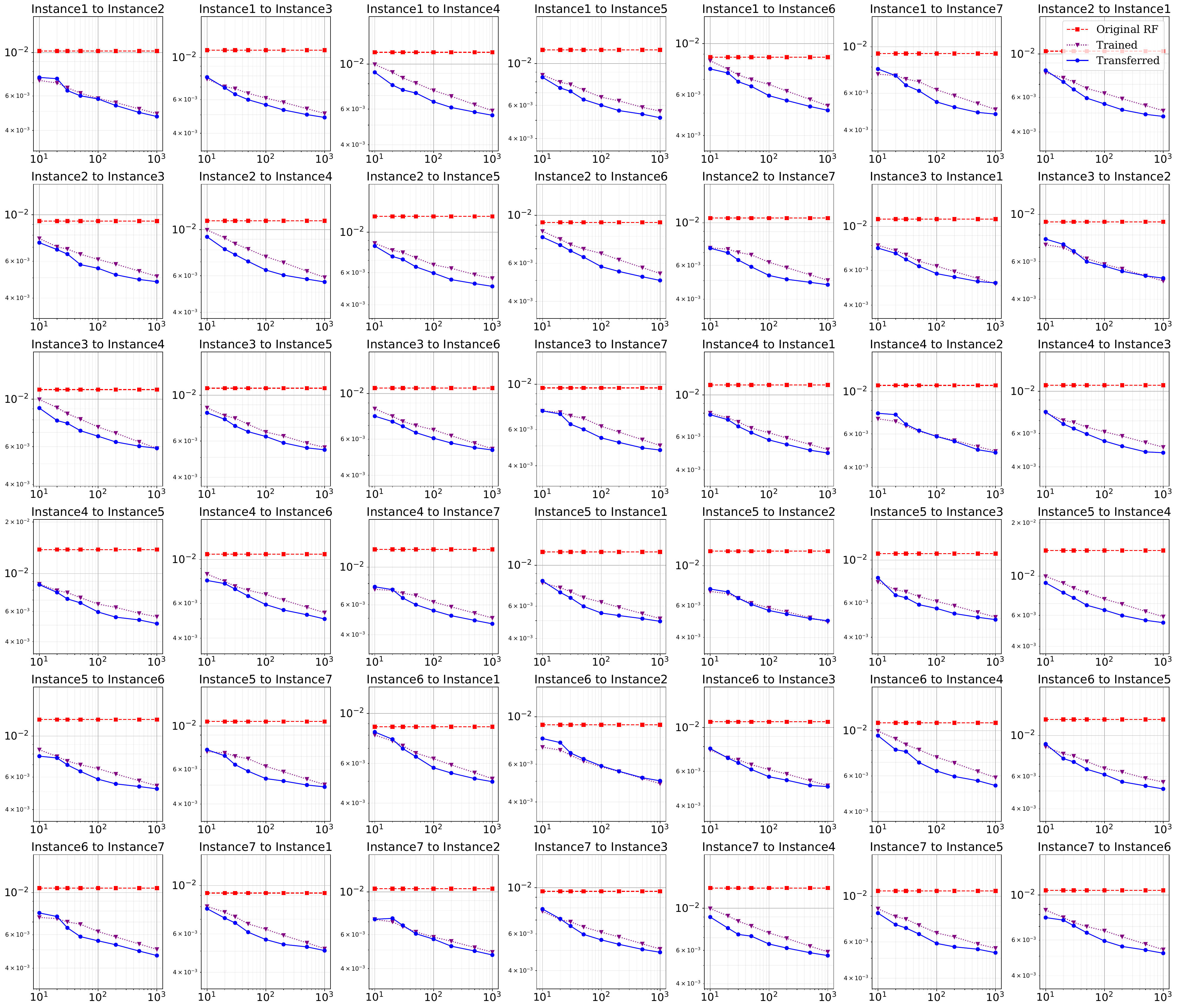}
  \caption{The SMAPE values (displayed on the $y$-axis) for three model variants—original RFR, transferred RFR, and a model trained exclusively on the transfer dataset—are analyzed on F5 of Single-Objective Game-Benchmark MarioGAN Suite. These values are plotted against transfer dataset sizes (shown on the $x$-axis), which include 10, 20, 30, 50, 100, 200, 500, and $1\,000$ samples.
  }
  \label{figure:RFRSMAPEplot_Real_world_decoration_frequency_overworld}
\end{figure*}

\begin{figure*}[htbp]
  \centering
  \includegraphics[width=\textwidth]{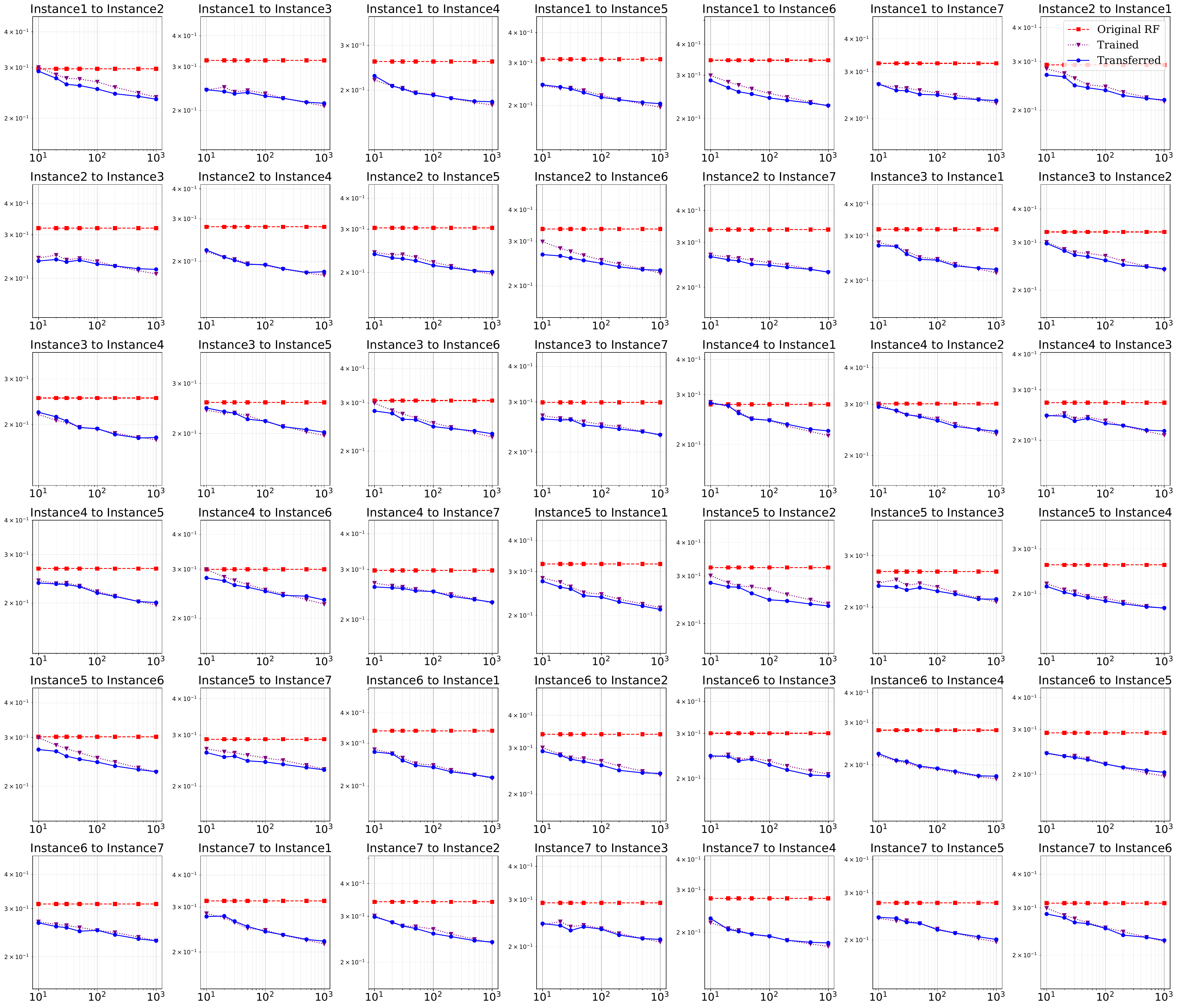}
  \caption{The SMAPE values (displayed on the $y$-axis) for three model variants—original RFR, transferred RFR, and a model trained exclusively on the transfer dataset—are analyzed on F1 of Single-Objective Game-Benchmark MarioGAN Suite. These values are plotted against transfer dataset sizes (shown on the $x$-axis), which include 10, 20, 30, 50, 100, 200, 500, and $1\,000$ samples.
  }
  \label{figure:RFRSMAPEplot_enemy_distribution_underground}
\end{figure*}

\end{document}